\documentclass{article}

\PassOptionsToPackage{numbers, sort&compress}{natbib}

\usepackage[preprint]{neurips_2021}

\usepackage[utf8]{inputenc} %
\usepackage[T1]{fontenc}    %
\usepackage{booktabs}       %
\usepackage{amsfonts}       %
\usepackage{nicefrac}       %
\usepackage{microtype}      %
\usepackage[backref=page]{hyperref}

\usepackage{amsmath, amssymb,amsthm}
\usepackage{pifont}
\usepackage{graphicx}
\usepackage{bbm}
\usepackage{subcaption}
\usepackage[font=small]{caption}

\usepackage{todonotes}
\usepackage{enumitem}
\usepackage{float}
\usepackage{booktabs}
\usepackage{colortbl}
\usepackage{xcolor}
\usepackage{multirow}

\usepackage{titletoc}
\usepackage{hyperref}
\usepackage[page,header]{appendix}
\usepackage[T1]{fontenc}    %
\usepackage{url}            %
\usepackage{booktabs}       %
\usepackage{amsfonts}       %
\usepackage{nicefrac}       %
\usepackage{microtype}      %
\usepackage{hyperref}
\usepackage{graphicx}
\usepackage{algorithm2e}

\usepackage{mathrsfs}
\usepackage{amssymb}
\usepackage{graphicx,wrapfig,lipsum}
\usepackage{caption}
\usepackage{float}
\usepackage{subcaption}
\usepackage{enumitem}
\usepackage{siunitx}
\usepackage{multirow}
\usepackage{tikz}
\usepackage{footnote}

\usepackage{lipsum}
\usepackage{frame}
\usepackage[framemethod=TikZ]{mdframed} 
\usepackage[multiple]{footmisc}
\usepackage{placeins}

\usepackage{longtable}

\usepackage{url}

\definecolor{mydarkblue}{rgb}{0,0.08,0.45}
\hypersetup{ %
    pdftitle={},
    pdfauthor={},
    pdfsubject={},
    pdfkeywords={},
    pdfborder=0 0 0,
    pdfpagemode=UseNone,
    colorlinks=true,
    linkcolor=mydarkblue,
    citecolor=mydarkblue,
    filecolor=mydarkblue,
    urlcolor=mydarkblue,
    pdfview=FitH}

\usepackage{listings}
\usepackage{color}
\definecolor{codegreen}{rgb}{0,0.5,0}
\definecolor{codeblue}{rgb}{0,0,0.9}
\definecolor{codegray}{rgb}{0.5,0.5,0.5}
\definecolor{codepurple}{rgb}{0.58,0,0.82}
\definecolor{backcolour}{rgb}{0.95,0.95,0.92}
\definecolor{backcolour2}{rgb}{0.9,0.9,0.9}
\definecolor{codered}{rgb}{0.5,0,0}
\definecolor{textcodered}{rgb}{0.4,0,0}
\definecolor{palegray}{rgb}{0.98,0.98,0.99}

\lstdefinestyle{mystyle}{
    backgroundcolor=\color{backcolour},
    commentstyle=\color{codegreen},
    keywordstyle=\color{codeblue},
    numberstyle=\tiny\color{codegray},
    stringstyle=\color{codegreen},
    breakatwhitespace=false,
    breaklines=true,
    captionpos=b,
    keepspaces=true,
    numbersep=5pt,
    showspaces=false,
    showstringspaces=false,
    showtabs=false,
    tabsize=2,
    otherkeywords={with},
    basicstyle=\ttfamily\scriptsize
}

\definecolor{Gray}{gray}{0.90}
\definecolor{White}{RGB}{255,255,255}
\newcolumntype{g}{>{\columncolor{Gray}}c}
\definecolor{ffe1da}{RGB}{255,225,218}
\definecolor{F7E0D5}{RGB}{247,224,213}
\definecolor{40E0D0}{RGB}{175,238,238}
\definecolor{darkF7E0D5}{RGB}{209,154,128}
\colorlet{Light}{backcolour}

\usetikzlibrary{backgrounds}
\usetikzlibrary{arrows,shapes}
\usetikzlibrary{tikzmark}
\usetikzlibrary{calc}

\usepackage{amsmath}
\usepackage{amsthm}
\usepackage{amssymb}
\usepackage{mathtools, nccmath}
\usepackage{wrapfig}
\usepackage{comment}

\usepackage{graphicx}

\usepackage{xspace}

\usepackage{array}
\usepackage{ragged2e}
\newcolumntype{P}[1]{>{\RaggedRight\hspace{0pt}}p{#1}}
\newcolumntype{X}[1]{>{\RaggedRight\hspace*{0pt}}p{#1}}

\usepackage{tcolorbox}

\usepackage{tikz}
\usetikzlibrary{arrows,shapes,positioning,shadows,trees,mindmap}
\usepackage[edges]{forest}
\usetikzlibrary{arrows.meta}
\colorlet{linecol}{black!75}
\usepackage{xkcdcolors} %

\usepackage{tikz}
\usetikzlibrary{backgrounds}
\usetikzlibrary{arrows,shapes}
\usetikzlibrary{tikzmark}
\usetikzlibrary{calc}

\newcommand\ttt[1]{\ensuremath{\texttt{#1}}}

\newcommand{\e}{\begin{equation}}
\newcommand{\ee}{\end{equation}}

\title{DeXtreme: Transfer of Agile In-hand Manipulation from Simulation to Reality}

\author{
Ankur Handa\thanks{Equal contribution \newline \newline All authors are with NVIDIA (except for Denys Makoviichuk, who is with Snap). In addition to NVIDIA, Arthur Allshire, Jingzhou Liu, and Ritvik Singh are also with the University of Toronto, and Aleksei Petrenko is with the University of Southern California. \newline \newline Author contributions \hyperref[sec:contributions]{listed at the end of the paper.}}
\And
Arthur Allshire\footnotemark[1]
\And
Viktor Makoviychuk\footnotemark[1]
\And
Aleksei Petrenko\footnotemark[1]
\And 
Ritvik Singh\footnotemark[1]
\And
Jingzhou Liu\footnotemark[1]
\And
Denys Makoviichuk
\And
Karl Van Wyk
\And
Alexander Zhurkevich
\And
Balakumar Sundaralingam
\And
Yashraj Narang
\And 
Jean-Francois Lafleche
\And
Dieter Fox
\And
Gavriel State
\AND
\\
{\large NVIDIA}
}
\date{2022}

\begin{document}

\makeatletter
    \let\@oldmaketitle\@maketitle%
    \renewcommand{\@maketitle}{\@oldmaketitle
    \begin{minipage}[c]{\textwidth}
    \centering
    \vspace{-20pt}
    \includegraphics[width=0.245\linewidth]{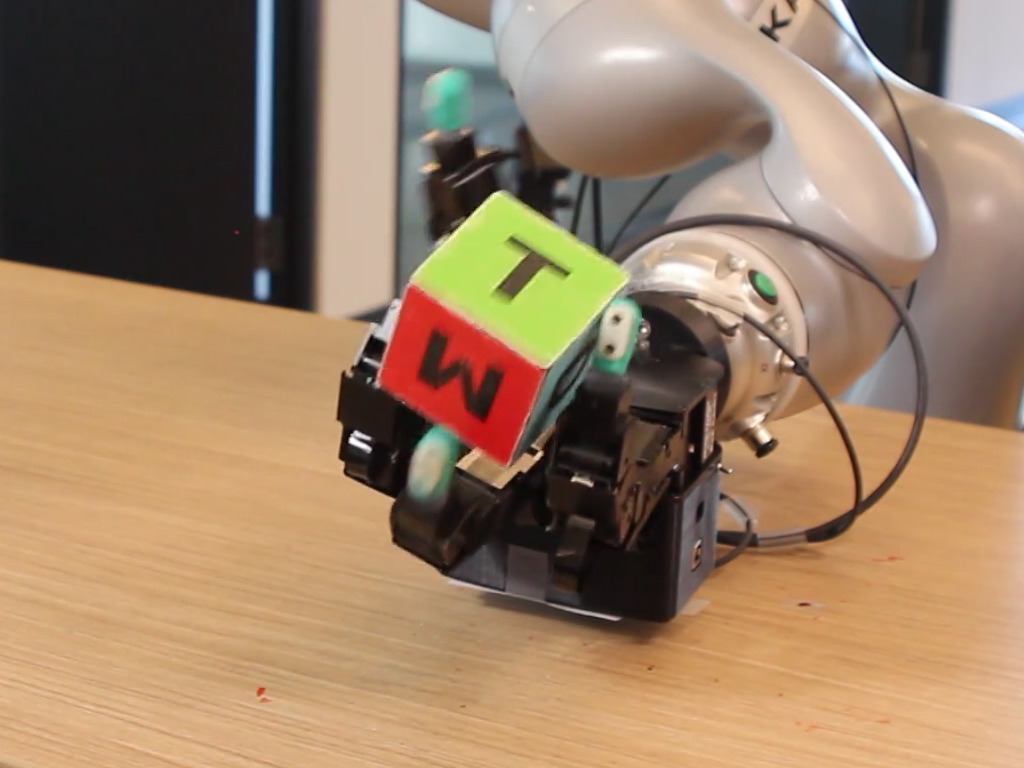}
    \includegraphics[width=0.245\linewidth]{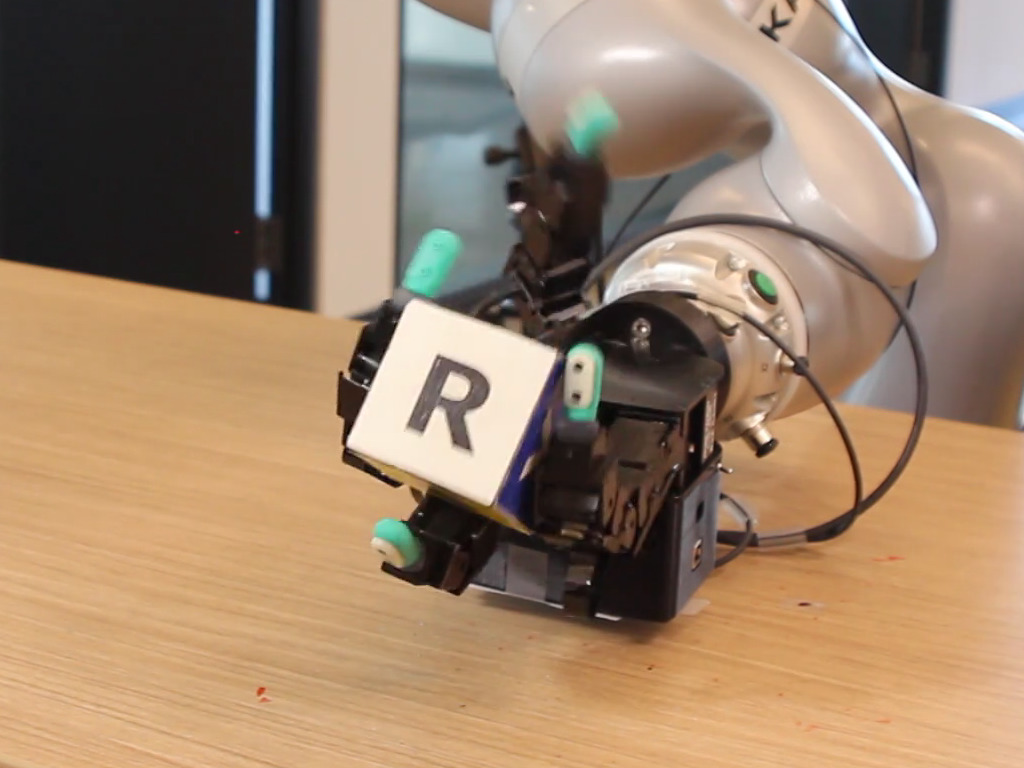}
    \includegraphics[width=0.245\linewidth]{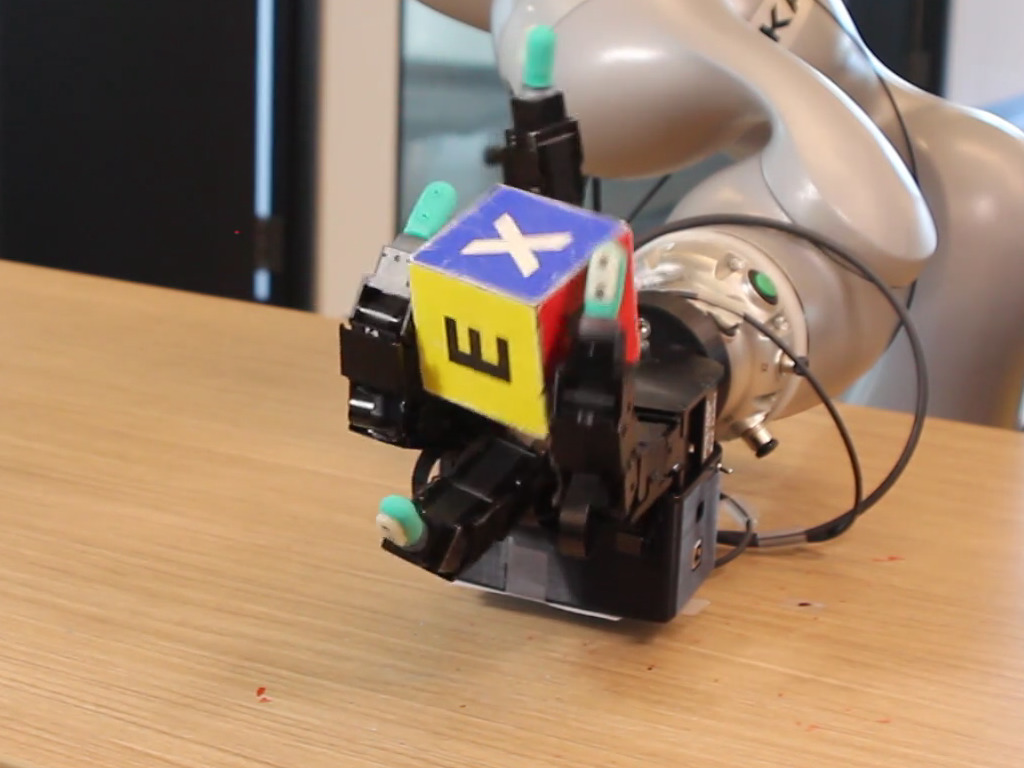}
    \includegraphics[width=0.245\linewidth]{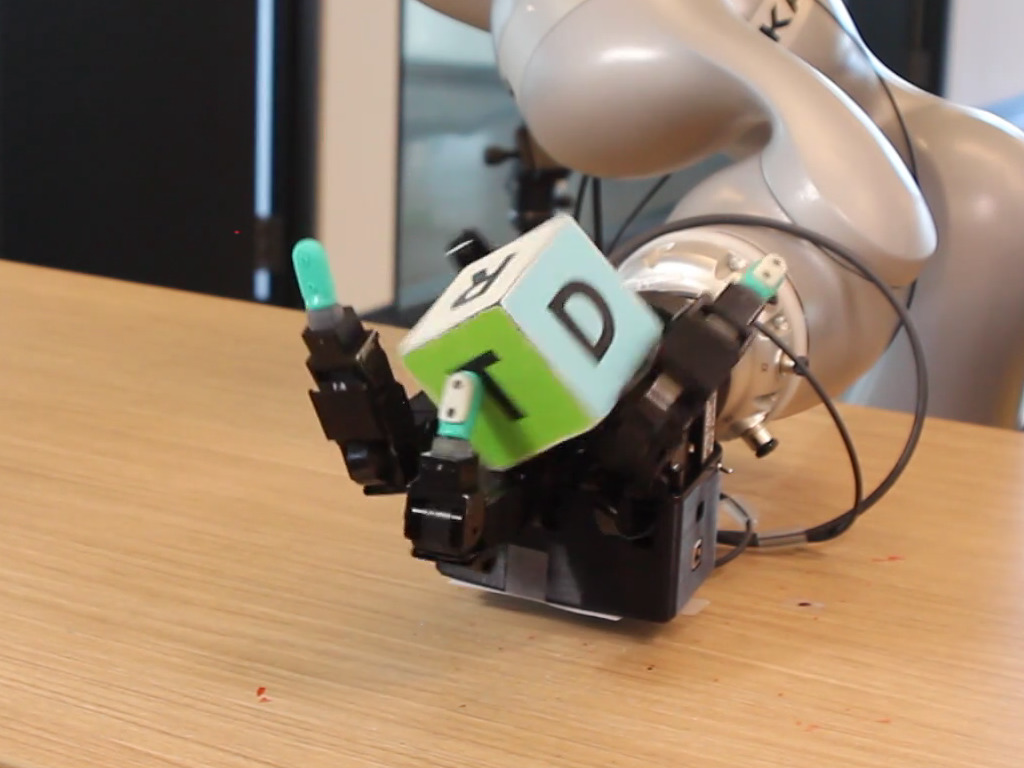}
    \captionof{figure}{The DeXtreme system using an Allegro Hand in action in the real world.}
    \label{fig:dextreme_header}
  \end{minipage}
  \vspace{2pt}
    }
\makeatother

\maketitle

\begin{abstract}
\vspace{-3mm}
Recent work has demonstrated the ability of deep reinforcement learning (RL) algorithms to learn complex robotic behaviours in simulation, including in the domain of multi-fingered manipulation. However, such models can be challenging to transfer to the real world due to the gap between simulation and reality. In this paper, we present our techniques to train a) a policy that can perform robust dexterous manipulation on an anthropomorphic robot hand and b) a robust pose estimator suitable for providing reliable real-time information on the state of the object being manipulated. Our policies are trained to adapt to a wide range of conditions in simulation. Consequently, our vision-based policies significantly outperform the best vision policies in the literature on the same reorientation task and are competitive with policies that are given privileged state information via motion capture systems. Our work reaffirms the possibilities of sim-to-real transfer for dexterous manipulation in diverse kinds of hardware and simulator setups, and in our case, with the Allegro Hand and Isaac Gym GPU-based simulation. Furthermore, it opens up possibilities for researchers to achieve such results with commonly-available, affordable robot hands and cameras. Videos of the resulting policy and supplementary information, including experiments and demos, can be found at \url{https://dextreme.org/}.
\end{abstract}

\tableofcontents

\section{Introduction}
\label{sec:intro}

Multi-fingered robotic hands offer an exciting platform to develop and enable human-level dexterity. Not only do they provide kinematic redundancy for stable grasps, but they also enable the repertoire of skills needed to interact with a wide range of day-to-day objects. However, controlling such high-DoF end-effectors has remained challenging. Even in research, most robotic systems today use parallel-jaw grippers. 

In 2018, \citet{openai-sh} showed for the first time that multi-fingered hands with a purely end-to-end deep-RL based approach could endow robots with unprecedented capabilities for challenging contact-rich in-hand manipulation. However, due to the complexity of their training architecture, and the \emph{sui generis} nature of their work on sim-to-real transfer, reproducing and building upon their success has proven to be a challenge for the community. Recent advancements in in-hand manipulation with RL have made progress with multiple objects and an anthropomorphic hand \cite{Chen:etal:CoRL21}, but those results have only been in simulation.

While the NLP and computer vision communities have reproduced and extended the successes of large-scale models like GPT-3~\cite{GPT:etal:arXiv20} and DALL-E \cite{DALLE:etal:arXiv21,DALLE2:arXiv:2022} respectively, similar efforts have remained elusive in robotics due to hardware and infrastructure challenges. Using large-scale data from simulations may provide avenues to unlock a similar step function in robotics capabilities.

This paper builds on top of the prior work in \cite{openai-sh}. We use a comparatively affordable Allegro Hand with a locked wrist and four fingers, using only position encoders on servo motors; the Shadow Hand used in OpenAI's experiments costs an order of magnitude more than the Allegro Hand. We also develop a simple vision system that requires no specialised tracking or infrastructure on the hand; the system works on three off-the-shelf RGB cameras compared to OpenAI's expensive marker-based setup, making our system easily accessible for everyone. Furthermore, we use the GPU-based Isaac Gym physics simulator \cite{isaacgym} as opposed to the CPU-based %
MuJoCo \cite{MuJoCo:etal:2012}, which allows us to reduce the amount of computational resources used and the complexity of the training infrastructure. Our best models required only 8 NVIDIA A40 GPUs to train, as opposed to OpenAI's use of a CPU cluster composed of 400 servers with 32 CPU-cores each, as well as 32 NVIDIA V100 GPUs \cite{openai-rubiks} (compute requirements for block reorientation). Our more affordable hand, in combination with the simple vision system architecture and accessible compute, dramatically simplifies the process of developing and deploying agile and dexterous manipulation. We summarise our contributions below:

\begin{itemize}
    \item We demonstrate a system for learning-based dexterous in-hand manipulation that uses low-cost hardware (one order of magnitude less expensive than \cite{openai-sh}), uses a purely vision-based pipeline, sets more diverse pose targets, uses orders-of-magnitude cheaper compute, and offers further insights into this problem with detailed ablations.
    \item We develop a highly robust pose estimator trained entirely in simulation which works through heavy occlusions and in a variety of robotic settings \textit{e.g.} \url{https://www.youtube.com/watch?v=-MTsm0Uh_5o}.
    \item While not directly comparable to \cite{openai-sh} due to different hardware, our purely vision-based state estimation results not only outperform their best vision-based results, but also fare comparably to their marker-based results.
    \item We will also release both our vision and RL pipelines for reproducibility. %
    We seek to provide a much broader segment of the research community with access to a novel state-of-the-art in-hand manipulation system in hopes of catalyzing further studies and advances. 
\end{itemize}

\section{Method}
\label{sec:method}

\subsection{Task}
\label{sec:task}

We propose a method for performing object reorientation on an anthropomorphic hand. Initially the object to be manipulated is placed on the palm of the hand and a random target orientation is sampled in $SO(3)$\footnote{In contrast to previous works \cite{openai-sh, openai-rubiks}, which limited it to configurations with flat faces pointing upwards.}. The policy then orchestrates the motion of the fingers so as to bring the object to its desired target orientation.
Similar to \citet{openai-sh}, if the object orientation is within a specified threshold of 0.4 radians of the target orientation, we sample a new target orientation. The fingers continue from the current configuration and aim to move the object to its new target orientation. The success criterion is the number of consecutive target orientations achieved without dropping the object or having the object stuck in the same configuration for more than 80 seconds. Importantly, each consecutive success becomes increasingly harder to achieve as the fingers have to keep the object in the hand without dropping, hence testing the policy's ability to model the dynamics on the go. 

\begin{figure}[htp]
\centering

\begin{subfigure}{\textwidth}
\centering
\includegraphics[width=0.7\linewidth]{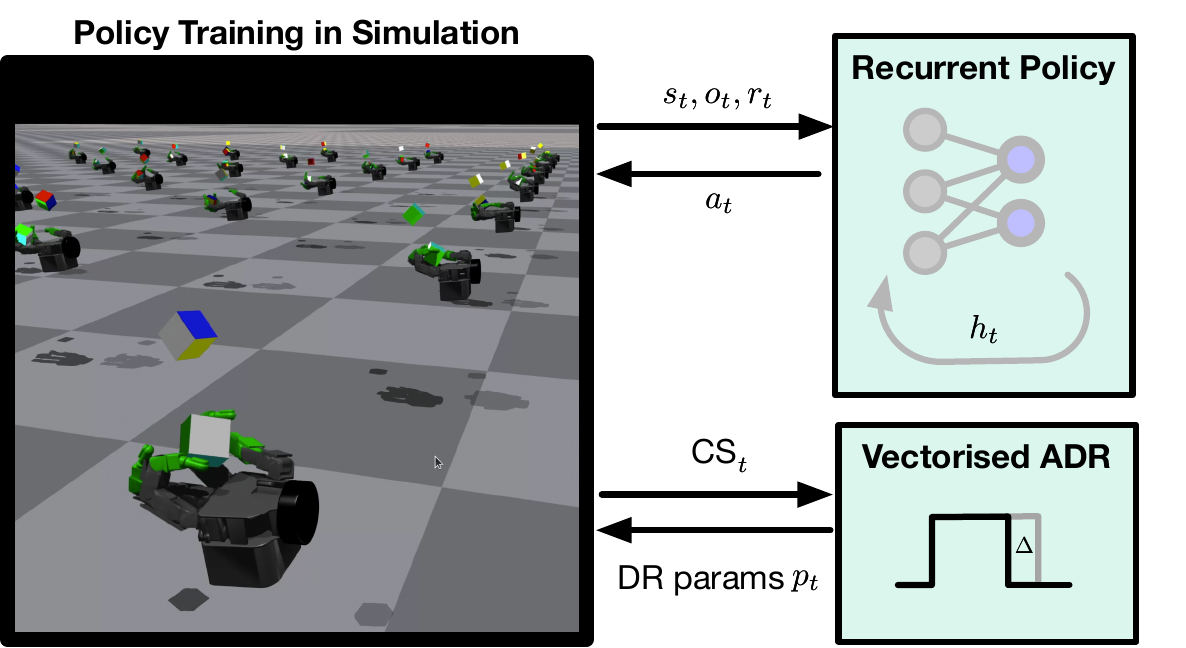}
\caption{Policy training.}
\end{subfigure}

\bigskip

\begin{subfigure}{\textwidth}
\centering
\includegraphics[width=0.8\linewidth]{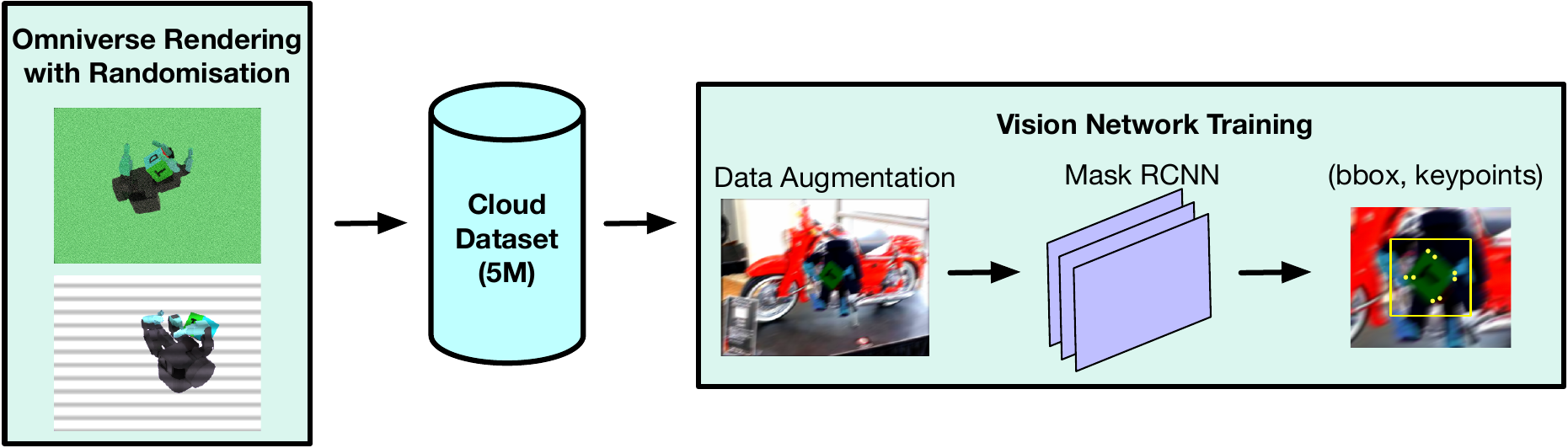}
\caption{Vision data generation and training pipeline.}
\end{subfigure}

\bigskip

\begin{subfigure}{\textwidth}
\centering
\includegraphics[width=0.7\linewidth]{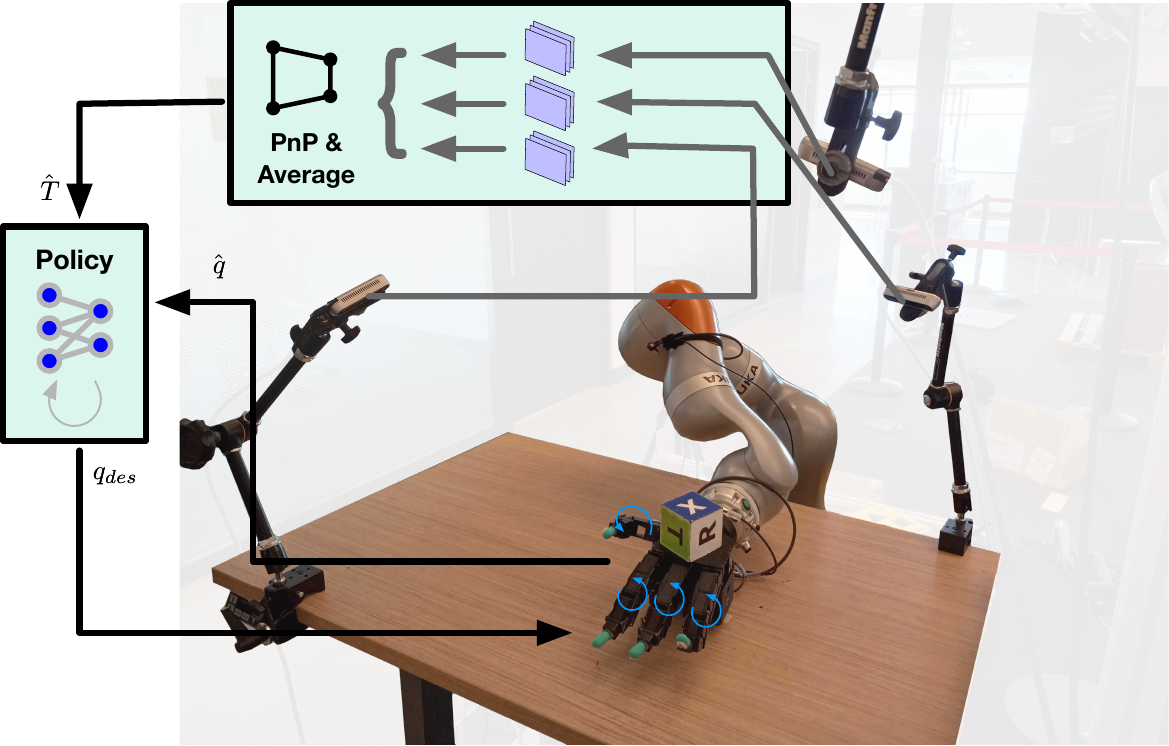}
\caption{Functioning in the real world.}
\end{subfigure}

\caption{High level overview of the training and inference systems.}
\label{fig:system}
\end{figure}

For a quick and high level understanding of this work, we encourage readers to watch the video \url{https://www.youtube.com/watch?v=TAUiaYAVkfI
}. Figure \ref{fig:system} provides a quick overview of different components involved in the system.

\subsection{Hardware}
\label{sec:hardware}

\begin{figure}[t]
    \centering
    \includegraphics[width=0.7\linewidth]{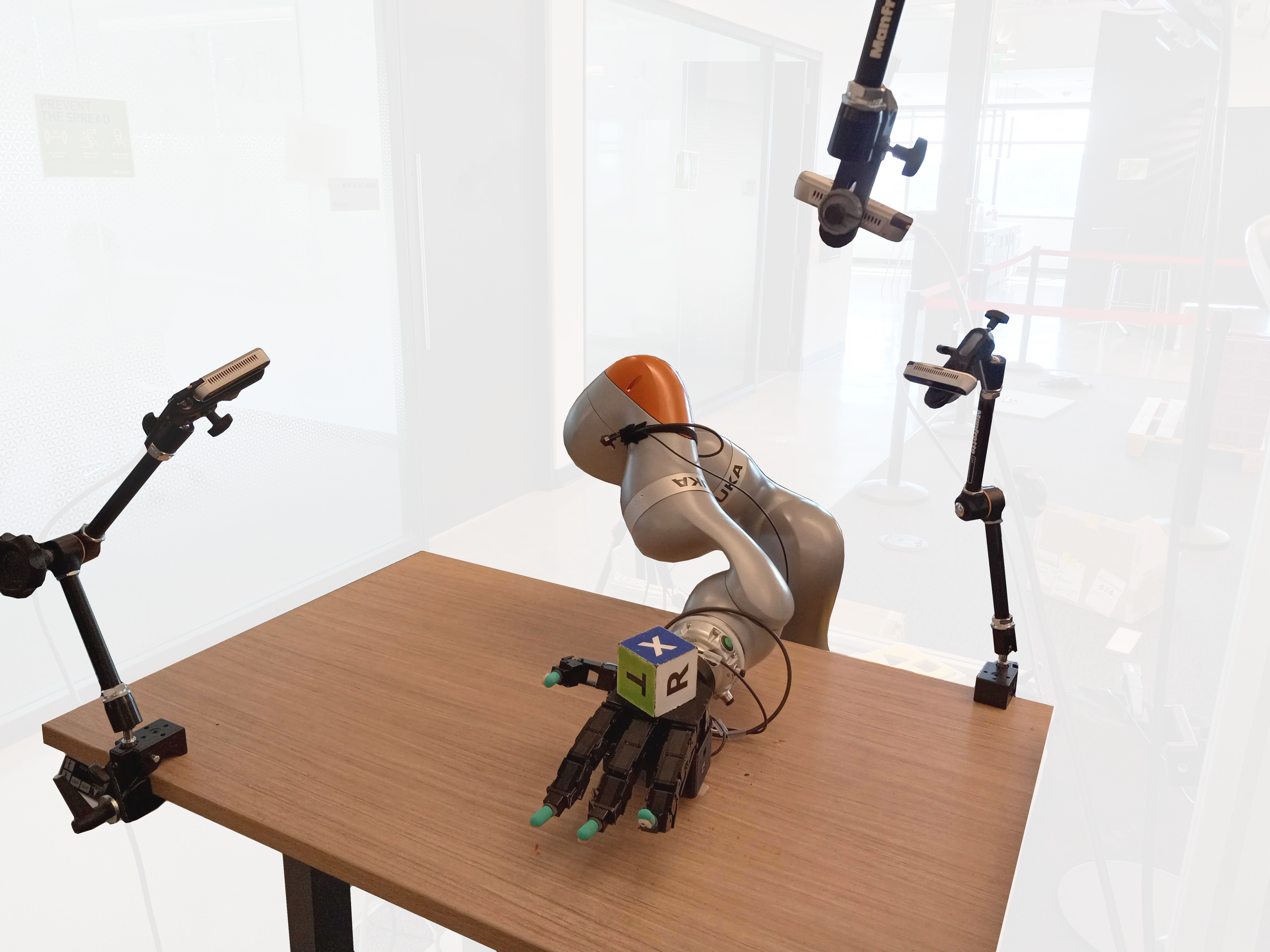}
    \caption{The hardware setup used in this work, unlike \cite{openai-sh}, is not housed in a cage, and our system is robust enough to perform the task in an open laboratory environment. The background in the image is alpha-blended for visibility.
    }   
    \label{fig:hardware}
\end{figure}

Our hardware setup (see Fig \ref{fig:hardware}) consists of an Allegro Hand rigidly mounted at the wrist. We use 3 Intel D415 cameras for object tracking with RGB frames \textit{i.e.} no depth images were used. The cameras are extrinsically calibrated relative to the palm link of the hand. Our object tracking is done entirely using a vision-based system, and in contrast to \cite{openai-sh}, we do not use any marker-based system to track the cube or fingertip states.

The object we learn to manipulate is a $6.5~\text{cm}$ cuboid with coloured and lettered stickers on the sides. These stickers allow the vision system to distinguish between different faces (see Sec. \ref{sec:pose-estimation}). The pose of the cube is represented with respect to the palm of the robot hand. The camera-camera extrinsics and camera-robot calibration allow us to transform the cube pose from the canonical reference frame of a camera to the palm. Since the cube is represented locally in the palm reference frame, the policy performance is not dependent on the physical location of the setup, enabling us to move the setup freely whenever desired.

\subsection{Policy Learning with RL}
\label{sec:policy-learning}

\textbf{RL Formulation}: The task of manipulating the cube to the desired orientation is modelled as a sequential decision making problem where the agent interacts with the environment in order to maximise the sum of discounted rewards. In our case, we formulate it as a discrete-time, partially observable Markov Decision Process (POMDP). We use Proximal Policy Optimisation (PPO) \cite{schulman2017proximal} to learn a parametric stochastic policy $\pi_\theta$  (actor), mapping from observations $o \in \mathcal{O}$ to actions $a \in \mathcal{A}$. PPO additionally learns a function $V^{\pi}_\phi(s, o)$ (critic) to approximate the on-policy value function. Following \citet{asymmetric-ac}, the critic does not take in the same observations as the actor, but receives additional observations including states $s \in \mathcal{S}$ in the POMDP. The actor and critic observations are detailed in Table \ref{table:policy-inputs}.

\begin{table}[!ht]
        \centering
        \resizebox{0.7\linewidth}{!}{%
        \centering
        \begin{tabular}{@{}llcc@{}}
        \toprule
        \rowcolor[HTML]{D4F7EE}
        \textbf{Input} & \textbf{Dimensionality} & \textbf{Actor} & \textbf{Critic} \\ \hline
        Object position with noise & 3D & \checkmark & \checkmark \\
        Object orientation with noise & 4D (quaternion) & \checkmark & \checkmark \\
        Target position & 3D & \checkmark & \checkmark \\
        Target orientation & 4D (quaternion) & \checkmark & \checkmark \\
        Relative target orientation & 4D (quaternion) & \checkmark & \checkmark \\
        Last actions & 16D & \checkmark & \checkmark \\
        Hand joints angles & 16D & \checkmark & \checkmark \\
        \midrule
        Stochastic delays & 4D & $\times$ & \checkmark \\
        Fingertip positions & 12D & $\times$ & \checkmark \\
        Fingertip rotations & 16D (quaternions) & $\times$ & \checkmark \\
        Fingertip velocities & 24D & $\times$ & \checkmark \\
        Fingertip forces and torques & 24D & $\times$ & \checkmark \\
        Hand joints velocities & 16D & $\times$ & \checkmark \\
        Hand joints generalised forces & 16D & $\times$ & \checkmark \\
        
        object scale, mass, friction & 3D & $\times$ & \checkmark \\
        Object linear velocity & 3D & $\times$ & \checkmark \\
        Object angular velocity & 3D & $\times$ & \checkmark \\
        Object position & 3D & $\times$ & \checkmark \\
        Object rotation & 4D & $\times$ & \checkmark \\
        Random forces on object & 3D & $\times$ & \checkmark \\
        Domain randomisation params & 78D & $\times$ & \checkmark \\
        Gravity vector & 3D  & $\times$ & \checkmark \\
        Rotation distances & 2D  & $\times$ & \checkmark \\
        Hand scale & 1D  & $\times$ & \checkmark \\
        \bottomrule
        \end{tabular}
    }
     \vspace{2pt}
     \caption{Observations of the policy and value networks. The input vector is 50D in size for policy and 265D for the value function.}
     \label{table:policy-inputs}
\end{table}

We use a high-performance PPO implementation from rl-games \cite{rl-games2022} with the following hyper-parameters: discount factor $\gamma$=0.998 \footnote{We found that following \citep{openai-sh} and setting the higher discount of 0.998 (as opposed to 0.99 as used with MLPs \citep{isaacgym}) was essential to allowing us to train LSTMs.}, clipping $\epsilon$=0.2. While in some experiments the learning rate was updated adaptively based on a fixed KL threshold 0.016, our best result was obtained using linear scheduling of the learning rate for the policy (start value $lr=1e{-4}$) and a fixed learning rate for the value function ($lr=5e{-5}$). Our best policy $\pi_\theta: \mathcal O \times \mathcal{H} \to \mathcal A$ was a Long Short-Term Memory (LSTM) network \citep{lstm-schmidhuber} taking in environment observations $o$ and previous hidden state $h \in \mathcal H$. We use an LSTM backpropagation through time (BPTT) truncation length of 16. The LSTM has 1024 hidden units with layer normalization and is followed by 2 multilayer perceptron (MLP) layers with sizes 512 and ELU activation \cite{clevert2015fast}. The action space $\mathcal A$ of our policy is the PD controller target for each of the 16 joints on the robot hand. The value function LSTM layer has 2048 hidden units with layer normalization, followed by 2 MLP layers with 1024 and 512 units respectively with ELU activation. The output of the policy is low-pass filtered with an exponential moving average (EMA) smoothing factor. During training this factor is annealed from 0.2 to 0.15. Our best results in the real world were obtained with an EMA of 0.1, which provided a balance between agility and stability of the motion, preventing the hardware from breaking or motor cables from burning.

\subsection{Reward Formulation}
The reward formulation is inspired by the Shadow hand environment in Isaac Gym\cite{isaacgym}, and described and justified in Table \ref{tab:reward}.

\begin{table}[!htbp]
\vspace{-2pt}
        \centering
        \resizebox{0.8\linewidth}{!}{%
        \begin{tabular}{l|c|c|c} 
        
            \toprule
            \rowcolor[HTML]{D4F7EE}
            Reward & Formula & Weight & Justification \\
            \midrule
            Rotation Close to Goal & $1 / (\mathsf{d} + 0.1)$ & 1.0 & Shaped reward to bring cube close to goal\\
            Position Close to Fixed Target & $||\mathsf{p_{object}} - \mathsf{p_{goal}}||$ & -10.0 & Encourage the cube to stay in the hand \\
            Action Penalty & $||\mathsf{a}||^2$ & -0.001 & Prevent actions that are too large  \\
            Action Delta Penalty & $||\mathsf{targ_{curr}} - \mathsf{targ_{prev}}||^2$ & -0.25 & Prevent rapid changes in joint target \\
            Joint Velocity Penalty & $||\mathsf{v_{joints}}||^2 $& -0.003 & Stop fingers from moving too quickly\\
            \midrule
            \rowcolor[HTML]{D4F7EE}
            Reset Reward & Condition & Value & \\
            \midrule
            Reach Goal Bonus & $\mathsf{d} < 0.1$ & 250.0 & Large reward for getting the cube to the target \\
            \bottomrule
        \end{tabular}
    }
     \vspace{2pt}
     \caption{Reward terms are computed, multiplied by their weight, and summed to produce the reward at each timestep. $\mathsf{d}$ represents the rotational distance from the object's current to the target orientation. $\mathsf{p_{object}}$ and $\mathsf{p_{goal}}$ are the position of the object and goal respectively. $a$ is the current action. $\mathsf{targ_{curr}}$ and $\mathsf{targ_{prev}}$ are the current and previous joint position targets. %
     $\mathsf{v_{joints}}$ is the current joint velocity vector.}
     \label{tab:reward}
\end{table}

\subsection{Simulation}
\label{sec:simulation}

\begin{table}[!t]
\centering
    \begin{subtable}[h]{0.9\linewidth}
        \resizebox{\linewidth}{!}{%
        \centering
        \begin{tabular}{l|c|c|c|c} 
        
            \toprule
            \rowcolor[HTML]{D4F7EE}
            Parameter & Type & Distribution & Initial Range & ADR-Discovered Range \\
            \midrule
            
            \rowcolor[HTML]{EFEFEF} 

            \multicolumn{5}{l}{\textbf{Hand}} \\
            Mass & Scaling & $ \textrm{uniform}$ & [0.4, 1.5] & [0.4, 1.5] \\ %
            Scale & Scaling & $ \textrm{uniform}$ & [0.95, 1.05] & [0.95, 1.05] \\ %
            Friction & Scaling & $ \textrm{uniform}$ & [0.8, 1.2] & [0.54, 1.58]\\
            Armature & Scaling &  \textrm{uniform} & [0.8, 1.02] & [0.31, 1.24] \\ 
            Effort & Scaling & \textrm{uniform} & [0.9, 1.1] & [0.9, 2.49] \\ 
            Joint Stiffness & Scaling & $ \textrm{loguniform}$ & [0.3, 3.0] & [0.3, 3.52]\\
            Joint Damping & Scaling & $ \textrm{loguniform}$ & [0.75, 1.5] & [0.43, 1.6] \\
            Restitution & Additive & $ \textrm{uniform}$ & [0.0, 0.4] & [0.0, 0.4]\\

            \rowcolor[HTML]{EFEFEF} 
            \multicolumn{5}{l}{\textbf{Object}} \\
            Mass & Scaling & $ \textrm{uniform}$ & [0.4, 1.6] & [0.4, 1.6] \\
            Friction & Scaling & $ \textrm{uniform}$ & [0.3, 0.9] & [0.01, 1.60]\\
            Scale & Scaling & $ \textrm{uniform}$ & [0.95, 1.05] & [0.95, 1.05]\\
            External Forces & Additive & Refer to~\cite[]{openai-sh} & -- & -- \\
            Restitution & Additive & $ \textrm{uniform}$ & [0.0, 0.4] & [0.0, 0.4] \\
            
            \rowcolor[HTML]{EFEFEF} 
            \multicolumn{5}{l}{\textbf{Observation}} \\
            Obj. Pose Delay Prob. & Set Value & $ \textrm{uniform}$ & [0.0, 0.05] & [0.0, 0.47] \\
            Obj. Pose Freq. & Set Value & $ \textrm{uniform}$ & [1.0, 1.0] & [1.0, 6.0]\\
            Obs Corr. Noise & Additive & $ \textrm{gaussian}$ & [0.0, 0.04] & [0.0, 0.12] \\
            Obs Uncorr. Noise & Additive & $ \textrm{gaussian}$ & [0.0, 0.04] & [0.0, 0.14] \\
            Random Pose Injection & Set Value & $ \textrm{uniform}$ & [0.3, 0.3] & [0.3, 0.3]\\

            \rowcolor[HTML]{EFEFEF} 
            \multicolumn{5}{l}{\textbf{Action}} \\
            Action Delay Prob. & Set Value & $ \textrm{uniform}$ & [0.0, 0.05] & [0.0, 0.31] \\
            Action Latency & Set Value & $ \textrm{uniform}$ & [0.0, 0.0] & [0.0, 1.5] \\
            Action Corr. Noise & Additive & $ \textrm{gaussian}$ & [0.0, 0.04] & [0.0, 0.32] \\
            Action Uncorr. Noise & Additive & $ \textrm{gaussian}$ & [0.0, 0.04] & [0.0, 0.48]\\
            RNA $\alpha$ & Set Value & \textrm{uniform} & [0.0, 0.0] & [0.0, 0.16] \\

            \rowcolor[HTML]{EFEFEF} 
            \multicolumn{5}{l}{\textbf{Environment}} \\
            Gravity (each coord.) & Additive & $ \textrm{normal}$ & [0, 0.5] & [0, 0.5] \\

            \bottomrule
            \end{tabular}
        }
     \end{subtable}
     \vspace{2pt}
     \caption{\footnotesize Domain randomisation parameter ranges for policy learning}%
     \label{tab:dr}
     \vspace{-4mm}
\end{table}

Our aim in this paper is to learn dexterous manipulation behaviours. Current on-policy learning algorithms can struggle to accomplish this on real robots due to the number of samples required. Hence, we learn our behaviours entirely in simulation. We use the GPU-based Isaac Gym physics simulator \citep{isaacgym}, which models contacts differently than MuJoCo's soft-contact model \cite{MuJoCo:etal:2012} used in \cite{openai-sh}. Isaac Gym gives the advantage of being able to simulate thousands of robots in parallel on a single GPU, mitigating the need for large amounts of CPU resources.

\subsection{Domain Randomisation}

It is widely known that there is a "sim-to-real" gap between physics simulators and real life \cite{RealityGap:1995}. Compounding this is the fact that the robot as a system can change from day to day (\textit{e.g.}, due to wear-and-tear) and even from timestep to timestep (\textit{e.g.}, stochastic noise). To help overcome this, we introduce various kinds of randomisations \cite{Peng:etal:ICRA2018} into the simulated environment
as listed in Table \ref{tab:dr}.

\textbf{Vectorised Automatic Domain Randomisation}: In our best policies, we set the parameters of the domain randomisations via a vectorised implementation of Automatic Domain Randomisation (ADR, introduced in \citep{openai-rubiks}). ADR automatically adjusts the range of domain randomisations to keep them as wide as possible while keeping the policy performance above a certain threshold. This allows us to train policies with less randomisation earlier in training (enabling behaviour exploration) while producing final policies that are robust to the largest range of environment conditions possible at the end of training, with the aim of improving sim-to-real transfer by learning policies which are robust and adaptive to a range of environment randomisation parameters. Using the parallelisation provided by Isaac Gym, we implement a vectorised variant of the algorithm, which we call Vectorised Automatic Domain Randomisation (VADR, see Algorithm \ref{alg:vadr}).

The range of randomisations for each environment parameter in ADR is modelled as a uniform distribution $d^n \sim U(p^{2n}, p^{2n+1})$, where $p^{2n}$ and $p^{2n+1}$ are the current lower and upper randomisation boundaries, and $n \in {0, \dots D-1}$ is the parameter index for each of the $D$ ADR dimensions. Each dimension starts with initial values from system calibration or best-guesses for the randomisation bounds, $p^{2n}_{init}$ and $p^{2n+1}_{init}$ for the lower and upper bounds of parameter $n$, respectively. Optional minimum and maximum bounds on the randomisations may also be specified, $p^{n}_{min}$ and $p^{n}_{max}$.
Unlike in \cite{openai-rubiks}, we choose the size of step $\Delta^n$ separately for each parameter. This trades off more tuning work for more stable training and the mitigation of the need for custom, secondary distributions on top of certain randomisation dimensions, as were used in that work.

\begin{figure}[htp]
\includegraphics[width=0.45\linewidth]{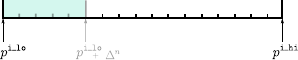}
\includegraphics[width=0.45\linewidth]{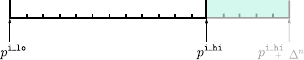}
\centerline{
\hfill
\makebox[0.5\linewidth][c] { \footnotesize{(a) $mean(Q^{\texttt{i\_lo}}) < t_l$}}
\hfill
\makebox[0.5\linewidth][c] { \footnotesize{(b) $mean(Q^{\texttt{i\_hi}}) > t_h$}}
}
\\

\includegraphics[width=0.45\linewidth]{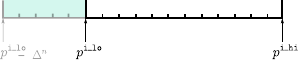}
\includegraphics[width=0.45\linewidth]{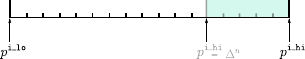}
\vspace{2mm}
\centerline{
\hfill
\makebox[0.5\linewidth][c] { \footnotesize{(c) $mean(Q^{\texttt{i\_lo}}) > t_h$}}
\hfill
\makebox[0.5\linewidth][c] { \footnotesize{(d) $mean(Q^{\texttt{i\_hi}}) < t_l$}}
}
\caption{\small
Parameter range adjustments, $p^{\texttt{i\_lo}}$ and $p^{\texttt{i\_hi}}$, with ADR based on the performance of policy at the boundaries $Q^{\texttt{i\_lo}}$ and $Q^{\texttt{i\_hi}}$ with respect to the thresholds $t_l$ and $t_h$. If the mean performance at the boundary $Q^{\texttt{i\_lo}}$ is less than threshold $t_l$, the range is tightened (a) and if it is above a threshold $t_h$, the range is expanded (c). Similarly, if the mean performance at the boundary $Q^{\texttt{i\_hi}}$ is above a threshold $t_h$, the range is expanded (b) and if it is lower than $t_l$, the range is tightened (d).} \vspace{1mm}
\label{fig:adr_range_adjustments}
\end{figure}

Evaluation proceeds as follows: environments sample a value for each randomisation dimension uniformly between the upper and lower bounds. A fraction (40\%) of the vectorised environments are dedicated to evaluation. In these environments, one of the ADR randomisation dimensions is fixed to the current lower or upper boundary (the rest of the dimensions are sampled from the aforementioned uniform distribution set by ADR). The episode proceeds to roll out; the number of consecutive successes is recorded at the end of the episode. This figure is added to a queue for the boundary of maximum length $N=256$. Then, if mean consecutive sucesses on the boundary is above a certain threshold, $t_H=20$, the range is widened, and if the performance is below a lower threshold $t_L=5$, then the range is tightened on that bound. This is depicted in Figure \ref{fig:adr_range_adjustments}. If on a particular step the value of a bound changes, the queue is cleared (as the previous performance data then becomes invalid). In this way, ADR will discover the maximum ranges over which a policy can perform, including for example discovering the limits of parameters impacting physics stability. Our vectorised ADR (VADR) full algorithm is described below in Algorithm \ref{alg:vadr} and Algorithm \ref{alg:training-loop}.

\RestyleAlgo{ruled}
\SetKwComment{Comment}{/* }{ */}
\SetKwInput{KwInit}{Init}
\SetKwInput{KwParams}{Parameters}

\SetKwProg{Fn}{Function}{}{}

\begin{algorithm}
\caption{Vectorisied Automatic Domain Randomisation (VADR)}\label{alg:vadr}

\Fn{ADRUpdate (p, \texttt{success\_counts}, \texttt{ADR\_mode})}{
\tcc{Function to decide whether to update each ADR boundary's value based on the success level of just-completed episodes.}
    \For{$n=0, \dots, D-1$} {
        $\ttt{i\_lo} = 2n$ \;
        $\ttt{i\_hi} = 2n+1$ \;
        $Q^{\ttt{i\_lo}} \gets \texttt{success\_counts[ADR\_modes==i\_lo]}$\;
        $Q^{\ttt{i\_hi}} \gets \texttt{success\_counts[ADR\_modes==i\_hi]}$\;
        \uIf{length($Q^{\ttt{i\_lo}}$)$==N$}{\
            \uIf{mean($Q^{\ttt{i\_lo}}$) $>t_h$}{
                $p^{\ttt{i\_lo}} \gets p^{\ttt{i\_lo}} - \Delta^n$
            }
            \uElseIf{mean($Q^{\ttt{i\_lo}}$) $<t_l$} {
                $p^{\ttt{i\_lo}} \gets p^{\ttt{i\_lo}} + \Delta^n$
            }
            \uIf{$p^\ttt{i\_lo}$ changed}{
                $p^\ttt{i\_hi} \gets clip(p^\ttt{i\_lo}, {p^n_{min}}, {p^n_{max}})$ \;
                $Clear(Q^{\ttt{i\_lo}})$
            }
        } 
        
        \uIf{length($Q^{\ttt{i\_hi}}$)$==N$}{\
            \uIf{mean($Q^{\ttt{i\_hi}}$) $>t_h$}{
                $p^{\ttt{i\_hi}} \gets p^{\ttt{i\_hi}} + \Delta^n$
            }
            \uElseIf{mean($Q^{\ttt{i\_hi}}$) $<t_l$} {
                $p^{\ttt{i\_hi}} \gets p^{\ttt{i\_hi}} - \Delta^n$
            }
            \uIf{$p^\ttt{i\_hi}$ changed}{
                $p^\ttt{i\_hi} \gets clip(p^\ttt{i\_hi}, {p^n_{min}}, {p^n_{max}})$ \;
                $Clear(Q^{\ttt{i\_hi}})$
            }
        }

    }
    \Return{p}
}

\Fn{ResetADRVals(p, \texttt{ADR\_modes}, \texttt{ADR\_vals})} {
\tcc{Function to resample ADR values for the next episode just after an episode is done.\\
NB. all operations here are vectorised, so eg. ADR\_modes is a tensor with 60\% of values -1.
}
    $\texttt{ADR\_modes} =
        \begin{cases}
            -1,& \text{with } p=60\%\\
            \sim \textbf{Categorical}(0, 2D-1), & \text{with } p=40\%
        \end{cases}
    $\;
    
    \tcc{Resample ADR values for each dimension for the next episode.}
    \For{$n=0, \dots, D-1$} {
        $\texttt{is\_eval} = \texttt{ADR\_modes == 2n | ADR\_modes == 2n+1}$ 
        $\texttt{ADR\_vals\{n\}} = 
        \begin{cases}
            p^\ttt{ADR\_modes}, & \text{where } \ttt{is\_eval}\\
            \sim U(p^{2n}, p^{2n+1}),              & \text{otherwise}
        \end{cases}$\;
    }
    \Return{\ttt{ADR\_modes}, \ttt{ADR\_vals}}
}
\end{algorithm}

\begin{algorithm}
Initialise tensor $\ttt{ADR\_modes} \in \mathbb Z^{N_{envs}}$ storing whether each environment is a normal environment (entry$=-1$) or performing sampling at one of the boundaries (entry$\in 0, \dots 2D-1$) \;
Initialise tensor $\ttt{ADR\_vals} \in R^{N_{envs}\times D}$ storing ADR randomisation values for each parameter in each environment (matrix storing values of $\phi$ from text).\;
Initialise ADR boundaries $p \in \mathbb R^{2D}$
\caption{Training with VADR.}\label{alg:training-loop}
\While{training} {
    \texttt{dones}, \texttt{successes} = \texttt{train\_step(\texttt{ADR\_vals})}\;
    $p$ = ADRUpdate(p, \texttt{successes[dones]}, \texttt{ADR\_mode[dones]}) \;
    \ttt{ADR\_modes[dones]}, \ttt{ADR\_vals[dones]} = ResetADRVals(p, \ttt{ADR\_modes[dones]}, \ttt{ADR\_vals[dones]}) \;
    
}

\end{algorithm}

When training on multiple GPUs (8 for our best policies), we ran VADR separately on each one. This was done for two reasons: firstly, to avoid additional synchronisation overhead of buffers. Secondly, to partially mitigate the disadvantage of ADR 
noted below in Section \ref{sec:quirks} 
caused by the failure to model the joint distribution; having multiple independent parameter sets to some extent will allow multiple sets of extreme parameters. All randomisations are set by ADR.\footnote{Except for mass and scale, which are randomised within a fixed range, as collision morphologies cannot be changed at run-time.} In the following, we describe the physics and non-physics randomisations in more detail.

\subsubsection{Physics Randomisations}
We apply physics randomisations to account for both changing real-world dynamics and the inevitable gaps between physics in simulation and reality. These include basic properties such as mass, friction and restitution of the hand and object. We also randomly scale the hand and object to avoid over-reliance on exact morphology. On the hand, joint stiffness, damping, and limits are randomised. Furthermore, we add random forces to the cube in a similar fashion to \cite{openai-sh}. 

\label{sec:physics-randomisation-details}

\textbf{Joint Stiffness, Joint Damping, and Effort} are scaled using the value sampled directly from the ADR-given uniform distribution.

\textbf{Mass and Scale} are randomised within a fixed range due to API limitations currently. However, we did not observe this as a significant limitation for our experiments, and our policies nevertheless achieved rollouts with high consecutive successes in the real world. 

\textbf{Gravity} cannot be randomised per-environment in Isaac Gym currently, but a new gravity value is sampled every $720$ concurrent simulation steps for all environments.

\subsubsection{Non-physics Randomisations}
In addition to normal physics randomisations, Table \ref{tab:dr} lists action and observation randomisations, which we found to be critical to achieving good real-world performance. To make our policies more robust to the changing inference frequency and jitter resulting from our ROS-based inference system, we add stochastic delays to cube pose and action delivery time as well as fixed-for-an-episode action latency. To the actions and observations, we add correlated and uncorrelated additive Gaussian noise. To account for unmodelled dynamics, we use a Random Network Adversary (RNA, see below). 

\label{sec:nonphysics-randomisation-details}

\textbf{Observation \& Action Noise}

We apply Gaussian noise to the observations and actions with the noise function

\begin{equation*}
f_{\mathsf{\delta}, \mathsf{\epsilon}}(x) = x + \delta + \epsilon 
\end{equation*}

Where $\delta$ and $\epsilon$ are sampled from Gaussian distributions \textit{parameterised by the ADR values $p^i, p^j$}, $\delta \sim \mathcal N(\cdot; 0, \text{var}(p^i))$, $\epsilon \sim \mathcal N(\cdot; 0, \text{var}(p^j))$
where
$\text{var}(a) = \exp\left[a^2\right]-1$

For $\delta$, this sampling happens once per episode at the beginning of the episode, corresponding to correlated noise. For $\epsilon$, sampling happens at every timestep. Note that the formula for $\text{var}$ has a cutoff at 0 noise. This allows ADR to set a certain fraction of environments to have 0 noise, which we found an important case that is not covered in previous works when setting fixed or above-zero cutoff variance (since during inference, zero white noise is added).

\textbf{Latency \& Delay}

We apply three forms of delay. The first is an exponential delay, where the chance of applying a delay each step is
$p^i$ and is given by $f(x; x_{last}) = x_{last} \cdot \mathsf{d} + x \cdot (1-\mathsf{d})$
and $\mathsf{d} \sim \text{Bern}(\cdot;p^i)$ is the Bernoulli distribution parametrised by the $i$-th ADR variable, $p^i \in [0, 1)$. This delay case, applied to both observations of cube pose and actions, mimics random jitter in latency times.

The second form of delay is action latency, where the action from n timesteps ago is executed. For this parameter, we slightly modify the vanilla ADR formulation to allow smooth increase in delay with ADR value despite the discretisation of timesteps. The bounds are still continuously modified, but the sampling from the range is done from a categorical distribution. Specifically, let $\epsilon \sim U(0, b) + U(-0.5, 0.5)$ be the sampled ADR value (plus random noise used to allow probabilistic blending of delay steps when sampling on the ADR boundary). Then the delay k is $k = \text{round}(\epsilon)$.

A third form of delay, this time on observation, is that caused by the refresh rate of the cameras in the real world. To compensate for this, we have randomisation on the refresh rate. Similarly to the aforementioned action latency, we use ADR to sample a categorical action delay $d \in \{1, \dots, delay_{max}\}$. We then only update the cube pose observation if $(t+r) \mod d = 0$, effectively mimicing a pose estimation frequency of $d \cdot \Delta t$  (where r is a randomly sampled alignment variable to offset updates from the beginning of the episode randomly).

\textbf{Random Pose Injection}

We noticed that due to heavy occlusion and caging from the fingers, our cube pose estimator exhibited occasional jumps. To ensure that the policy performance did not deteriorate and LSTM hidden state become corrupted by this, we occasionally inject completely random cube poses into the network. At the start of each episode for each environment, we sample a probability $p \in U(0, 0.3)$. Then each step we sample a variable $m \sim \text{Bern}(\cdot; p)$, and the cube pose becomes: $\mathsf{pose\_obs} = \mathsf{pose} \cdot (1-m) + \mathsf{random\_pose} \cdot m$.

\textbf{Random Network Adversary}

\begin{figure}[t]
    \centering
    \includegraphics[width=0.6\linewidth]{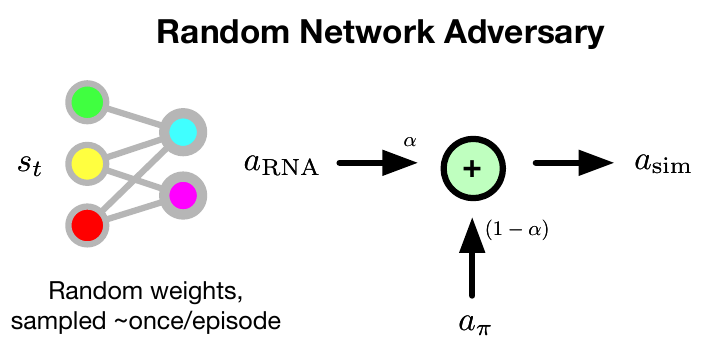}
    \caption{The functioning of the Random Network Adversary}   
    \label{fig:RNA}
\end{figure}

Random Network Adversary, introduced in \citep{openai-rubiks}, uses a randomly-generated neural network each episode to introduce much more structured, state-varying noise patterns into the environment, in contrast to normal Gaussian noise. As we are doing simulation on GPU rather than CPU, instead of using a new network per environment-episode and wasting memory on thousands of individual MLPs, we generate a single network across all environments and use a unique and periodically refreshed dropout pattern per environment. Actions from the RNA network are blended with those from the policy by $\mathbf{a} = \alpha \cdot \mathbf{a}_\text{RNA} + (1-\alpha) \cdot \mathbf{a}_{\text{policy}}$, where $\alpha$ is controlled by ADR. 

\subsubsection{Measuring ADR Performance in Training and in the Real World}

Nats per Dimension (npd) was a metric developed by OpenAI in \citep{openai-rubiks} to measure the amount of randomisation through the average entropy across the ADR distributions. While it does not directly capture the difficulty of the environment (since each dimension is not normalised for difficulty), it provides a rough proxy for how much randomisation there is in the environment. The formula for nats per dimension is given by:

\begin{equation}
    \text{npd} = \frac{1}{D} \sum_{n=0}^{D-1} \log\left(p^{2n+1} - p^{2n}\right)
\end{equation}

Currently, we measure real-world performance based on the number of consecutive successes. An avenue for future work is directly exploring how real-world policy performance corresponds to ADR randomisation levels in total and across different dimensions.

\subsection{Pose Estimation}
\label{sec:pose-estimation}

\textbf{Data Generation and Processing}: We use NVIDIA Omniverse Isaac Sim with Replicator\footnote{See \url{https://developer.nvidia.com/isaac-sim}} to generate 5M images of the cube in hand in just under a day. Each image is 320$\times$240 in resolution and contains visual domain randomisations as summarised in Table \ref{table:randomisations} to add variety to the training set. Such visual domain randomisations allow the network to be robust to different camera parameters and visual effects that may be present in the scene. In addition, we apply data augmentations during training on a batch in order to add even more variety to the training set. As such, a single rendered image from the dataset can provide multiple training examples with different data augmentation settings, thereby saving both rendering time and storage space. For instance, motion blur (important in our case where we have a fast-moving object to track) can be especially time-consuming at rendering time. Instead we generate it on the fly via motion-blur data augmentation by smearing the image with a Gaussian kernel with the blur direction and extent chosen randomly. The data augmentations used on the images are listed in Table \ref{table:data_aug_train}. Each augmentation is applied with a fixed probability to ensure that the batch consists of a combination of the original as well as augmented images.

\begin{table}[!t]
\centering
\begin{minipage}{.5\textwidth}
    \centering
    \resizebox{0.9\linewidth}{!}{%
    \centering
    \begin{tabular}{l|c} 
    
        \toprule
        \rowcolor[HTML]{D4F7EE}
        Parameter & Probability Distribution \\
        \midrule
        \rowcolor[HTML]{EFEFEF} 
        Albedo desaturation  & $ \textrm{uniform}(0.0, 0.3)$ \\
        Albedo add  & $ \textrm{uniform}(-0.2,0.2)$ \\
        \rowcolor[HTML]{EFEFEF} 
        Albedo brightness & $ \textrm{uniform}(0.5,1.0)$ \\
        Diffuse tint & $ \textrm{normal}(1.0, 0.3)$ \\
        \rowcolor[HTML]{EFEFEF} 
        Reflection roughness constant & $ \textrm{uniform}(0.0,1.0)$ \\
        Metallic constant & $ \textrm{uniform}(0.0,0.5)$ \\
        \rowcolor[HTML]{EFEFEF} 
        Specular level & $ \textrm{uniform}(0.0,1.0)$ \\ \hline
        \textbf{Enable camera noise} & $ \textrm{bernoulli}(0.3)$ \\
        \rowcolor[HTML]{EFEFEF} 
        \hspace*{0.4cm} Enable scan lines & $ \textrm{bernoulli}(0.3)$ \\
        \hspace*{0.8cm} Scan line spread & $ \textrm{uniform}(0.1, 0.5)$ \\
        \rowcolor[HTML]{EFEFEF} 
        \hspace*{0.4cm} Enable vertical lines & $ \textrm{bernoulli}(0.1)$ \\
        \hspace*{0.4cm} Enable film grain & $ \textrm{bernoulli}(0.2)$ \\
        \rowcolor[HTML]{EFEFEF} 
        \hspace*{0.8cm} Grain amount & $ \textrm{uniform}(0.0, 0.15)$ \\
        \hspace*{0.8cm} Grain size & $ \textrm{uniform}(0.7, 1.2)$ \\
        \rowcolor[HTML]{EFEFEF} 
        \hspace*{0.4cm} Enable random splotches & $ \textrm{bernoulli}(0.1)$ \\
        \hspace*{0.4cm} Colour amount & $ \textrm{uniform}(0.0, 0.3)$ \\
        \bottomrule
        \rowcolor[HTML]{EFEFEF} 
        Hand visibility & $ \textrm{bernoulli}(0.75)$ \\
        Camera pose & $ \textrm{Hemispherical shell}$\footnotemark  \\
        \rowcolor[HTML]{EFEFEF} 
        Camera f/l multiplier & $ \textrm{uniform}(0.99, 1.01)$ \\
        \bottomrule
    \end{tabular}
    }
    \vspace{2pt}
 \caption{Ranges of vision parameter randomisations.}
 \label{table:randomisations}
\end{minipage}%
\begin{minipage}{.5\textwidth}

\centering
\resizebox{0.8\linewidth}{!}{
\begin{tabular}{l|c} 
        \toprule
        \rowcolor[HTML]{D4F7EE}
        Data Augmentation Type & Probability \\
        \midrule
        \rowcolor[HTML]{EFEFEF} 
        CutMix (see \cite{cutmix19}) & 0.5 \\
        Random Blurring & 0.5 \\
        \rowcolor[HTML]{EFEFEF} 
        Random Background & 0.6 \\
        Random Rotation & 0.5 \\ 
        \rowcolor[HTML]{EFEFEF} 
        Random Brightness and Contrast & 0.5 \\
        Random Cropping and Resizing & 0.5 \\
        \rowcolor[HTML]{EFEFEF} 
        \bottomrule
\end{tabular}
}
\vspace{2pt}
\caption{Various data augmentations applied to the images on the fly during training. We also set a probability for each one of them.}
\label{table:data_aug_train}
\end{minipage}
\end{table}
\footnotetext{Camera pose is randomly sampled from a hemispherical shell with thickness 0.3m around the hand, with a random focus selected with a 0.2m radius around the hand origin.}

\begin{figure*}[t]
\centering
\begin{subfigure}{0.19\textwidth}
  \centering
    \includegraphics[width=\linewidth]{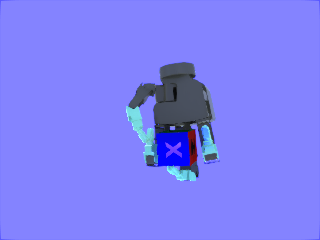}
\end{subfigure} %
\begin{subfigure}{0.19\textwidth}
  \centering
  \includegraphics[width=\linewidth]{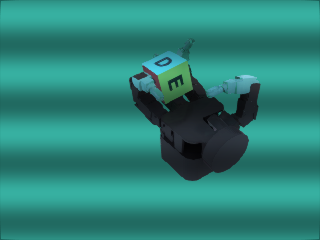}
\end{subfigure} %
\begin{subfigure}{0.19\textwidth}
  \centering
  \includegraphics[width=\linewidth]{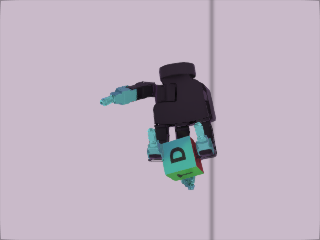}
\end{subfigure} %
\begin{subfigure}{0.19\textwidth}
  \centering
    \includegraphics[width=\linewidth]{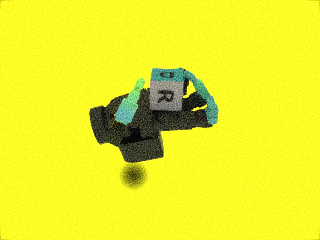}
\end{subfigure} %
\begin{subfigure}{0.19\textwidth}
  \centering
  \includegraphics[width=\linewidth]{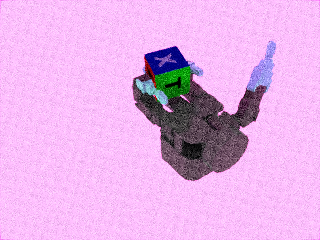}
\end{subfigure} %

\vspace{2pt}

\begin{subfigure}{0.19\textwidth}
  \centering
    \includegraphics[width=\linewidth]{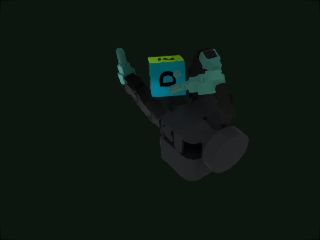}
\end{subfigure} %
\begin{subfigure}{0.19\textwidth}
  \centering
  \includegraphics[width=\linewidth]{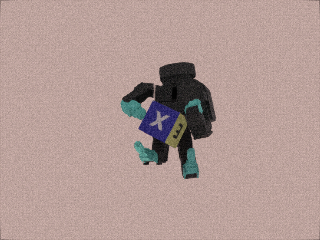}
\end{subfigure} %
\begin{subfigure}{0.19\textwidth}
  \centering
  \includegraphics[width=\linewidth]{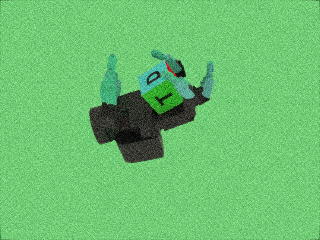}
\end{subfigure} %
\begin{subfigure}{0.19\textwidth}
  \centering
    \includegraphics[width=\linewidth]{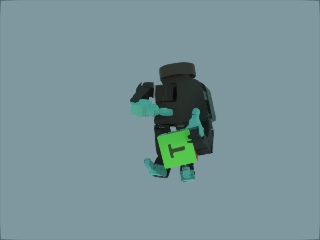}
\end{subfigure} %
\begin{subfigure}{0.19\textwidth}
  \centering
  \includegraphics[width=\linewidth]{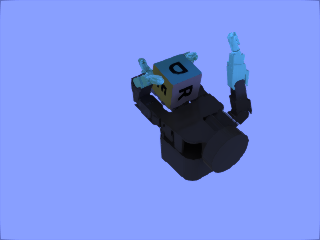}
\end{subfigure} %

\vspace{2pt}

\begin{subfigure}{0.19\textwidth}
  \centering
  \includegraphics[width=\linewidth]{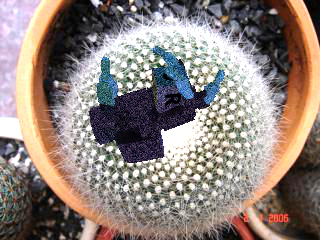}
\end{subfigure} %
\begin{subfigure}{0.19\textwidth}
  \centering
  \includegraphics[width=\linewidth]{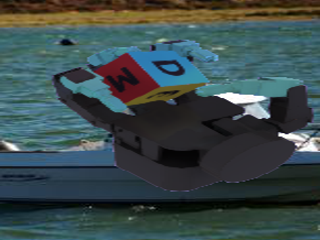}
\end{subfigure} %
\begin{subfigure}{0.19\textwidth}
  \centering
  \includegraphics[width=\linewidth]{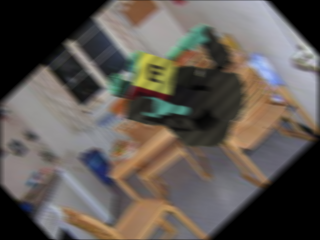}
\end{subfigure} %
\begin{subfigure}{0.19\textwidth}
  \centering
  \includegraphics[width=\linewidth]{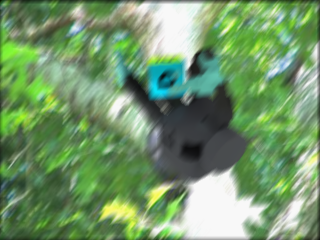}
\end{subfigure} %
\begin{subfigure}{0.19\textwidth}
  \centering
  \includegraphics[width=\linewidth]{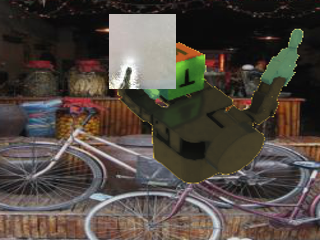}
\end{subfigure} %

\begin{subfigure}{0.19\textwidth}
  \centering
  \includegraphics[width=\linewidth]{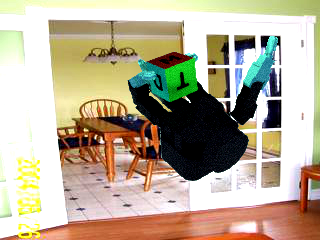}
\end{subfigure} %
\begin{subfigure}{0.19\textwidth}
  \centering
  \includegraphics[width=\linewidth]{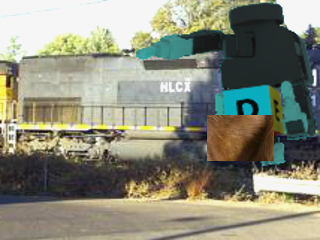}
\end{subfigure} %
\begin{subfigure}{0.19\textwidth}
  \centering
  \includegraphics[width=\linewidth]{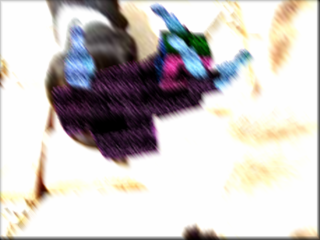}
\end{subfigure} %
\begin{subfigure}{0.19\textwidth}
  \centering
  \includegraphics[width=\linewidth]{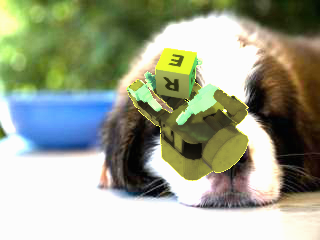}
\end{subfigure} %
\begin{subfigure}{0.19\textwidth}
  \centering
  \includegraphics[width=\linewidth]{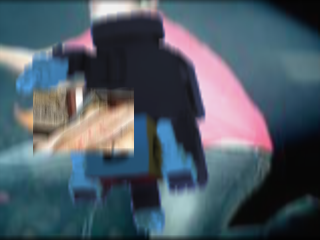}
\end{subfigure} %

\caption{Illustration of generated data. \textit{Top row:} Images generated from NVIDIA Isaac Sim, demonstrating different camera locations, lighting conditions, and camera poses. \textit{Bottom row:} Images after data augmentations such as CutMix, random cropping and rotation, and motion blur are applied during training.}
\end{figure*}

We also collect configurations of the cube in-hand generated by our policies running in the real world and play them back in Isaac Sim to render data where the pose estimates are not fully reliable \textit{i.e.} the pose estimator is not accurate all the time. This happens due to the sim-to-real gap as a result of either sparse sampling or insufficient lighting randomisations for those configurations. Playback in simulation enables dense sampling of pose and lighting around these configurations with more randomisations. This allows us to generate larger datasets that improve the reliability of the pose estimator in the real world \textit{i.e.} closing the perception loop between real and sim by mining configurations from the current best policy in the real world and using them in sim to render more images. We use the same intrinsics of the cameras in the real world, but randomise extrinsics when rendering data.

\textbf{Training setup and inference}: We use a torchvision Mask-RCNN \citep{maskrcnn}-inspired network \footnote{We also tried U-Net based direct regression of the keypoints from the image but found it to be somewhat unreliable. Sometimes the keypoints were detected at entirely different locations in the image where the cube was not even present. Having a network that first localises the cube and then detects the keypoints within the bounding box prevents such misdetections.} that regresses to the bounding box, segmentation, and keypoints located at the 8 corners of the cube. The bounding box localises the cube in the image, and the keypoint head regresses to the positions of the 8 corners of the cube within the bounding box. The networks are trained with cross-entropy loss for segmentation and keypoint location regression, and smooth L1 loss for bounding box regression. We use the ADAM optimiser with a learning rate of 1e-4. The network runs on three cameras at an inference rate of 20Hz on an NVIDIA RTX 3090 GPU and a 32-core AMD Ryzen Threadripper CPU. However, because the policy was trained with a control frequency of 30Hz in simulation, the pose estimator was locked to run at 15Hz to ensure that the policy receives pose observations at a constant integer interval of once every two control steps. To make the pose estimate reliable for the downstream policy, we first perform classic PnP \cite{PnP:2021} on each of the three cameras independently and then filter out the ones where the projected keypoints from the PnP pose do not match the inferred keypoints from the network. We triangulate the keypoints from the filtered cameras and register them against the model of the cube to obtain the pose in a canonical reference frame. We use the OpenCV implementation of PnP and roma \cite{roma:arxiv21} for registering keypoints against the model of the cube. We benchmark the pose on a test set consisting of 50K images and provide results in Table \ref{table:pose_results}. Since we do not use any marker-based system in the real world, we can only precisely evaluate the performance of the pose estimator in simulation. Our ablation studies in Section \ref{sec:real_world_perf} do test the strength of the pose estimator for manipulation in the real world. One important difference between our approach and \citet{openai-sh} is that our pose estimation is not done end-to-end. Since we detect keypoints in the image and use geometric computer vision to obtain the pose, our pose estimator is not tied to a fixed camera setup, unlike \cite{openai-sh}.

\begin{table}
\centering
\resizebox{0.7\linewidth}{!}{%
\begin{tabular}{l|l|lll}
\hline
\rowcolor[HTML]{D4F7EE}
Experiment & Avg. Rotation Error & \multicolumn{3}{c}{Avg. Translation Error} \\ \cline{3-5} 
\rowcolor[HTML]{D4F7EE}
 &  & \multicolumn{1}{c|}{X} & \multicolumn{1}{c|}{Y} & \multicolumn{1}{c}{Z} \\ \hline
\multicolumn{1}{c|}{Sim} & \multicolumn{1}{c|}{5.3$\pm$0.11$^{\circ}$} & \multicolumn{1}{c|}{1.9$\pm$0.1 mm} & \multicolumn{1}{c|}{4.1$\pm$0.2 mm} & \multicolumn{1}{c}{6.9$\pm$0.4 mm} \\ \hline
\end{tabular}
}
\vspace{2pt}
\caption{Rotation and translation error on test dataset with 90\% confidence intervals.}
\label{table:pose_results}
\end{table}

\section{Results}
\label{sec:results}

In the following section, we present the results we achieved in object reorientation in the simulations and then real world using the methods described in Section \ref{sec:method}. We then follow it up with tests of policy robustness in reality and simulation. 

\vspace{-3mm}
\subsection{Training in Simulation}
\begin{figure*}[h]
\vspace{-4mm}
\centering
\begin{subfigure}[c]{0.33\textwidth}
    \centering
    \includegraphics[width=0.98\textwidth]{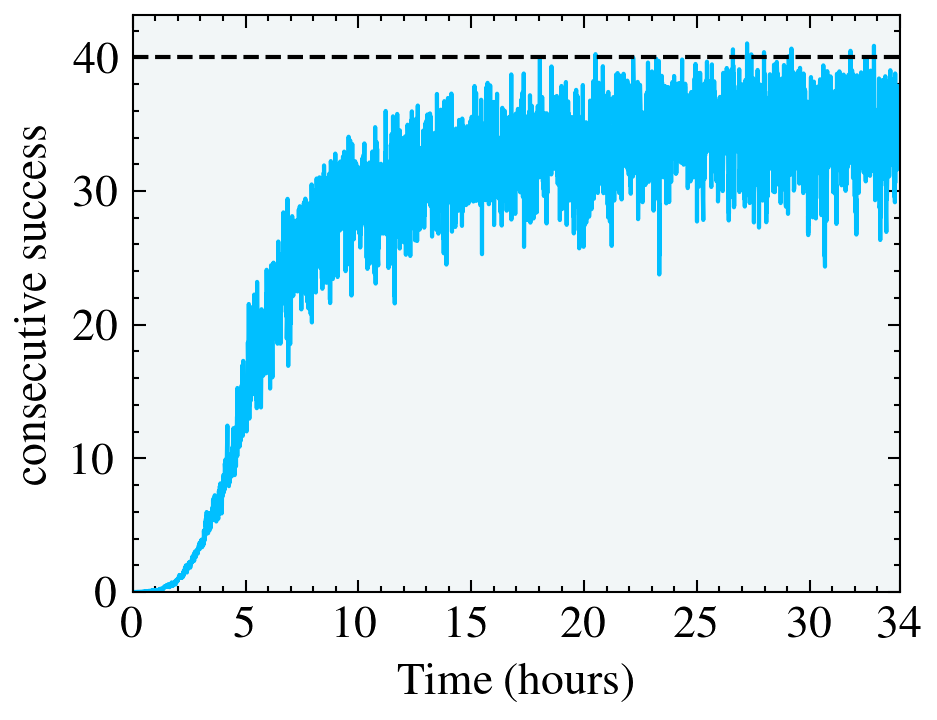}
    \caption{Manual DR training curve.}
\end{subfigure}%
\begin{subfigure}[c]{0.33\textwidth}
    \centering
    \includegraphics[width=0.98\textwidth]{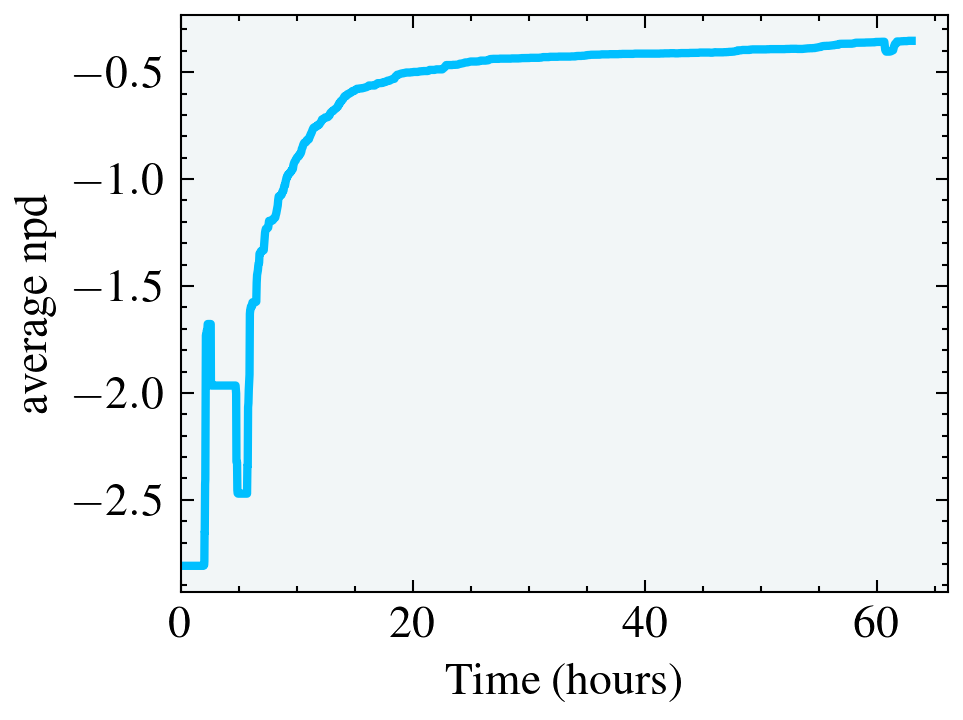}
    \caption{ADR npd evolution.}
\end{subfigure}%
\begin{subfigure}[c]{0.32\textwidth}
    \centering
    \includegraphics[width=0.98\textwidth]{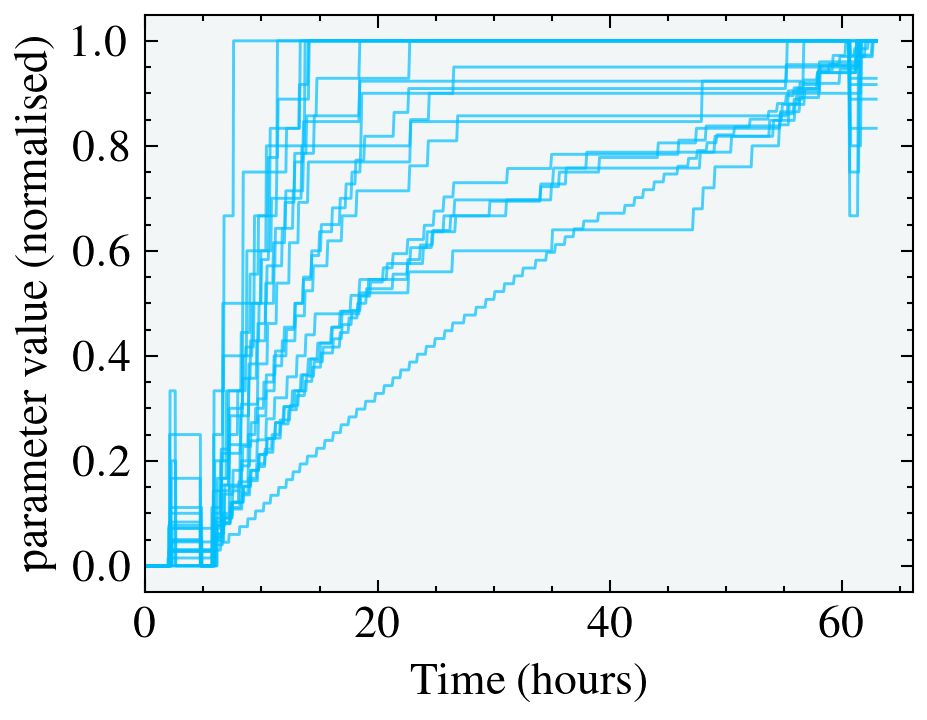}
    \caption{ADR parameter evolution.}
\end{subfigure}%
\caption{\footnotesize Training curve evolution for (a) manual domain randomisation, which takes $\sim$24 hours to reach an average of 35 consecutive successes, (b) ADR experiments showing how the average nats per distribution (npd), an indicator of the extent of randomisation, increases as a function of training (c) shows how parameters (normalised) optimised with ADR evolve over time. All curves are from experiments in simulations.}
\label{fig:training-curves}
\end{figure*}

\label{sec:training_in_sim}
For all of our experiments, we use a simulation $dt$ of $\frac{1}{60} s$ and a control $dt$ of $\frac{1}{30} s$. We train with 16384 agents per GPU and use a goal-reaching orientation threshold of 0.1 rad but test with 0.4 rad as in \cite{openai-sh, isaacgym} for all experiments both in simulation and the real world. All policies are trained with randomisations described in Table \ref{tab:dr}. %
Most importantly, our ADR policies using the same compute resources --- 8 NVIDIA A40s --- achieved the best performance in the real world after training for only 2.5 days in contrast to \citep{openai-rubiks} that trained for 2 weeks to months for the task of block reorientation\footnote{Although \citep{openai-rubiks} focused on the Rubik's cube, they also trained for block reorientation (\textit{pp.} 20, Table 3) with the same infrastructure to reproduce their results from \citep{openai-sh}.}. Various training curves for our task a) with manual DR, b) with automatic domain randomisation (ADR), and c) ADR parameter evolution are presented in Figure \ref{fig:training-curves}. We note that due to differences in physics engines and hand morphology, our simulation average consecutive successes are not directly comparable, but we achieve performance on par with \cite{openai-sh, openai-rubiks}. 

Training with manual DR takes roughly 32 hours to converge on 8 NVIDIA A40s generating a combined (across all GPUs) frame rate of 700K frames/sec. With a $dt=\frac{1}{60}$, this amounts to $\frac{32\times 700000}{60\times24\times365}$ which is $\sim$42 years of real-world experience.

\subsection{Real-World Policy Performance}
\label{sec:real_world_perf}

\begin{table*}[htp]
\centering
\resizebox{0.85\linewidth}{!}{
\begin{tabular}{c|c|c|c}
\hline 
\rowcolor[HTML]{D4F7EE}
\textbf{Experiment} & \textbf{Cons. Success Trials (sorted)} & \textbf{Average} & \textbf{Median} \\ \hline

\multirow{3}{*}{Best Model}                                       & 1, 6, 6, 10, 10, 18, 18, 36, 61, 112               & 27.8 $\pm$ 19.0 & 14.0\\ 
                                                          &   \cellcolor[HTML]{EFEFEF} 3, 4, 7, 16, 19, 22, 29, 31, 58, 77                & \cellcolor[HTML]{EFEFEF} 26.6 $\pm$ 13.2 & \cellcolor[HTML]{EFEFEF} 20.5 \\ 
                                                          & 1, 5, 5, 11, 12, 12, 33, 36, 42, 51 & 20.8 $\pm$ 9.8 & 12.0 \\ \hline

\multirow{3}{*}{\shortstack{Best Model \\ (Goal frame count=10)}}                                       & \cellcolor[HTML]{EFEFEF}  6, 8, 10, 16, 16, 17, 20, 33, 39, 45 & \cellcolor[HTML]{EFEFEF} 21.0  $\pm$ 7.4 & \cellcolor[HTML]{EFEFEF} 16.5

          \\ 
                                                          &   9, 11, 13, 13, 15, 16, 27, 29, 32, 36                & 20.1 $\pm$ 5.4 & 15.5                                                         \\ 
                                                          & \cellcolor[HTML]{EFEFEF}  2, 3, 3, 9, 11, 12, 14, 15, 43, 44               & \cellcolor[HTML]{EFEFEF} 16.6 $\pm$ 8.4 & \cellcolor[HTML]{EFEFEF} 11.5                                 \\ \hline

\multirow{3}{*}{Non-ADR Model}                                 & 2, 3, 7, 7, 13, 16, 22, 23, 26, 29                 & 14.8 $\pm$ 5.4 & 14.5                               \\ 
                                                          & \cellcolor[HTML]{EFEFEF}  1, 1, 3, 7, 8, 11, 14, 17, 22, 35                  & \cellcolor[HTML]{EFEFEF} 11.9 $\pm$ 5.8 & \cellcolor[HTML]{EFEFEF} 9.5                                 \\ 
                                                          & 0, 7, 8, 8, 9, 10, 10, 11, 17, 20                  & 10.0 $\pm$ 3.0 & 9.5                                 \\ \hline

\end{tabular}
}
\caption{The results of running different models on the real robot. We run 10 trials per policy \cite{openai-sh} to benchmark the average consecutive successes. Individual rows within each experiment indicate running the experiment on different days \cite{ABtesting:2021} and $\pm$ indicates 90\% confidence interval. Our best model was trained with ADR while non-ADR experiments had DR ranges manually tuned. The second experiment shows results when the cube is held at a goal for additional consecutive frames once the target cube pose is reached.}
\label{table:real_results}
\end{table*}

It is worth noting that, while in simulations, state information is derived directly from physics buffers, in all real-world experiments we use the pose estimator described in Section \ref{sec:pose-estimation} to obtain the pose of the cube and provide it as input to the policy. The qualitative results of this are best illustrated by the accompanying videos at \url{https://dextreme.org}. 

\begin{figure}[ht]
\centerline{
\includegraphics[width=0.8\linewidth]
{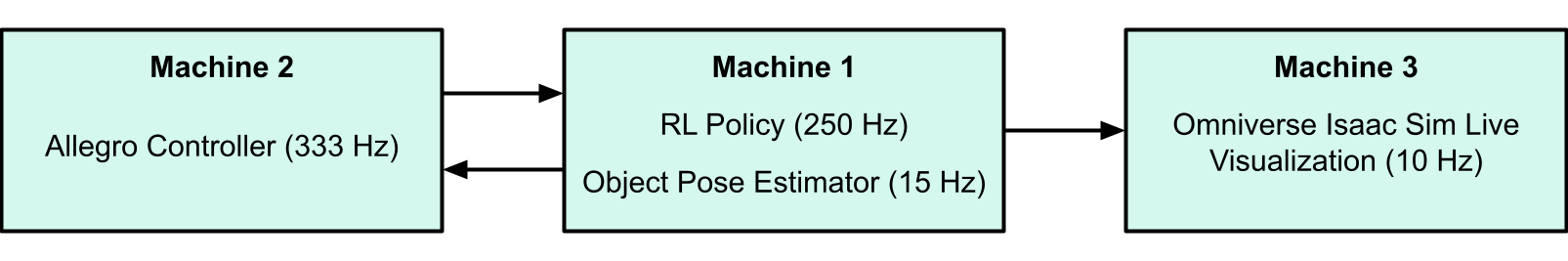}
}
\caption{\small
The system runs on a combination of three machines. The pose estimation and policy are run on the same machine, whereas the Allegro control and live Omniverse visualisation are done on two separate machines that communicate with the main machine at different rates via ROS messages.} \vspace{1mm}
\label{fig:deployment_pipeline}
\end{figure}

The deployment pipeline of the system is shown in Figure \ref{fig:deployment_pipeline}. We use three separate machines to run various components. The main machine has an NVIDIA RTX 3090, which runs both the policy as well as the pose estimator. We also do live visualisation of the real-world manipulation in Omniverse but disable the physics. 

Similar to \citep{openai-sh}, we observe a large range of different behaviours in the policies deployed on the real robot. Our real-world quantitative results measuring average consecutive successes are illustrated in Table \ref{table:real_results}. We collect 10 trials for each policy to obtain the average consecutive successes and also 
collect different sets of trials across different days to understand the inter-day variability that may arise due to different dynamics, temperature, and lighting conditions. We believe such inter-day variations are important to benchmark in robotics \cite{ABtesting:2021} and have endeavoured to highlight this specifically in this challenging task. We find that our policies do not show a dramatic drop in average performance, indicating that they are mostly robust to inter-day variations. 

\begin{figure}[htp]
\includegraphics[width=0.245\linewidth]{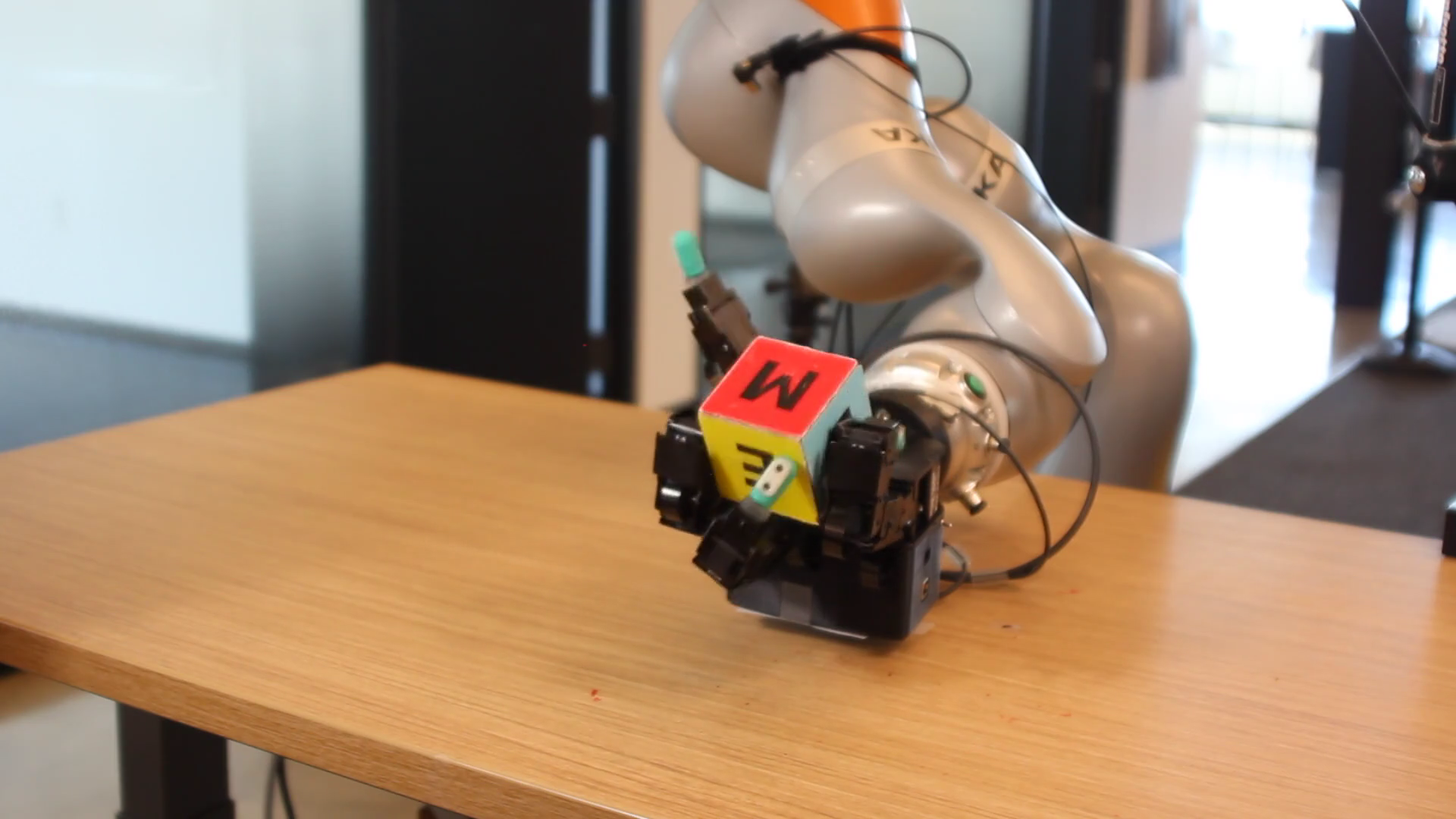}
\includegraphics[width=0.245\linewidth]{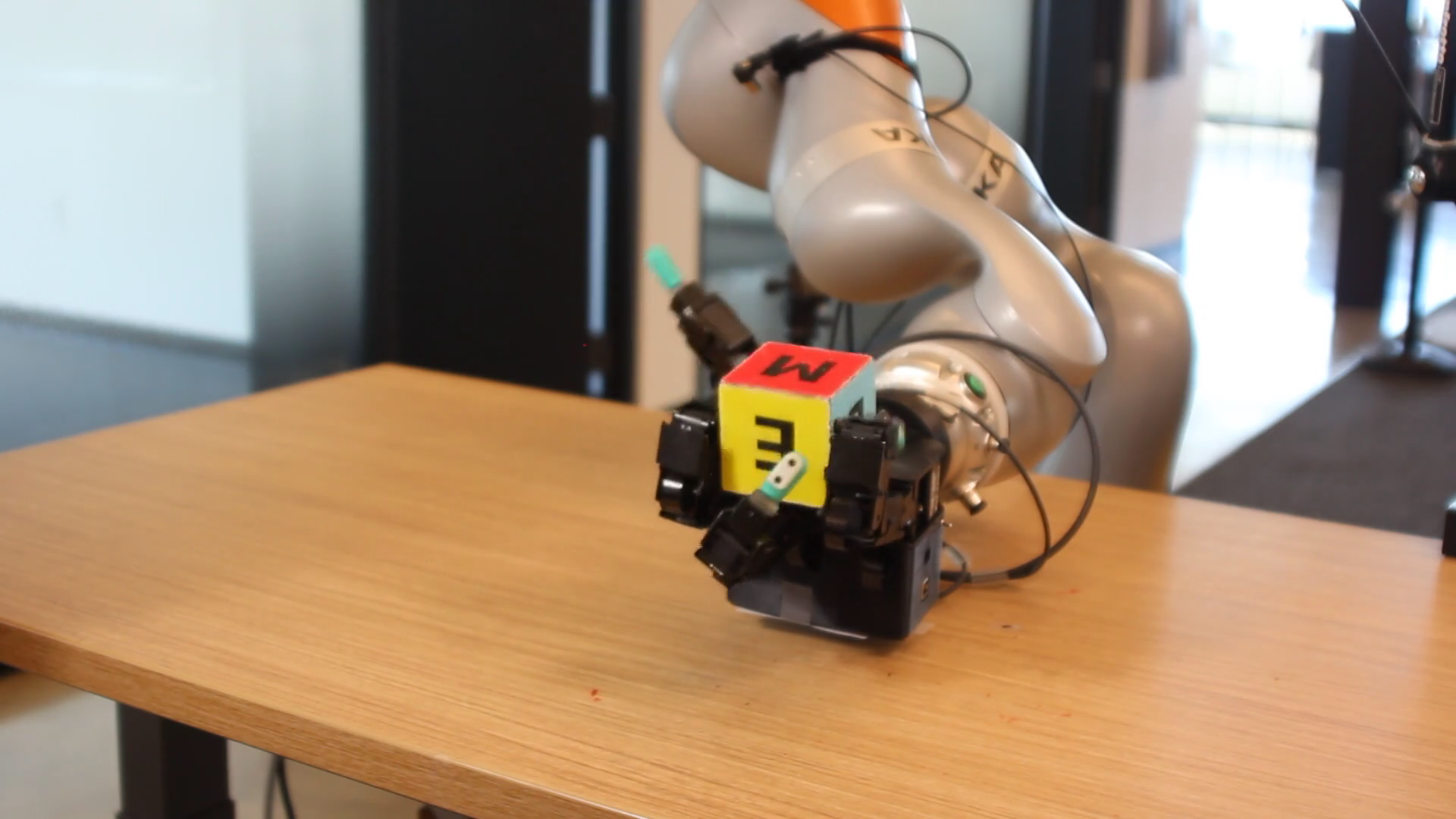}
\includegraphics[width=0.245\linewidth]{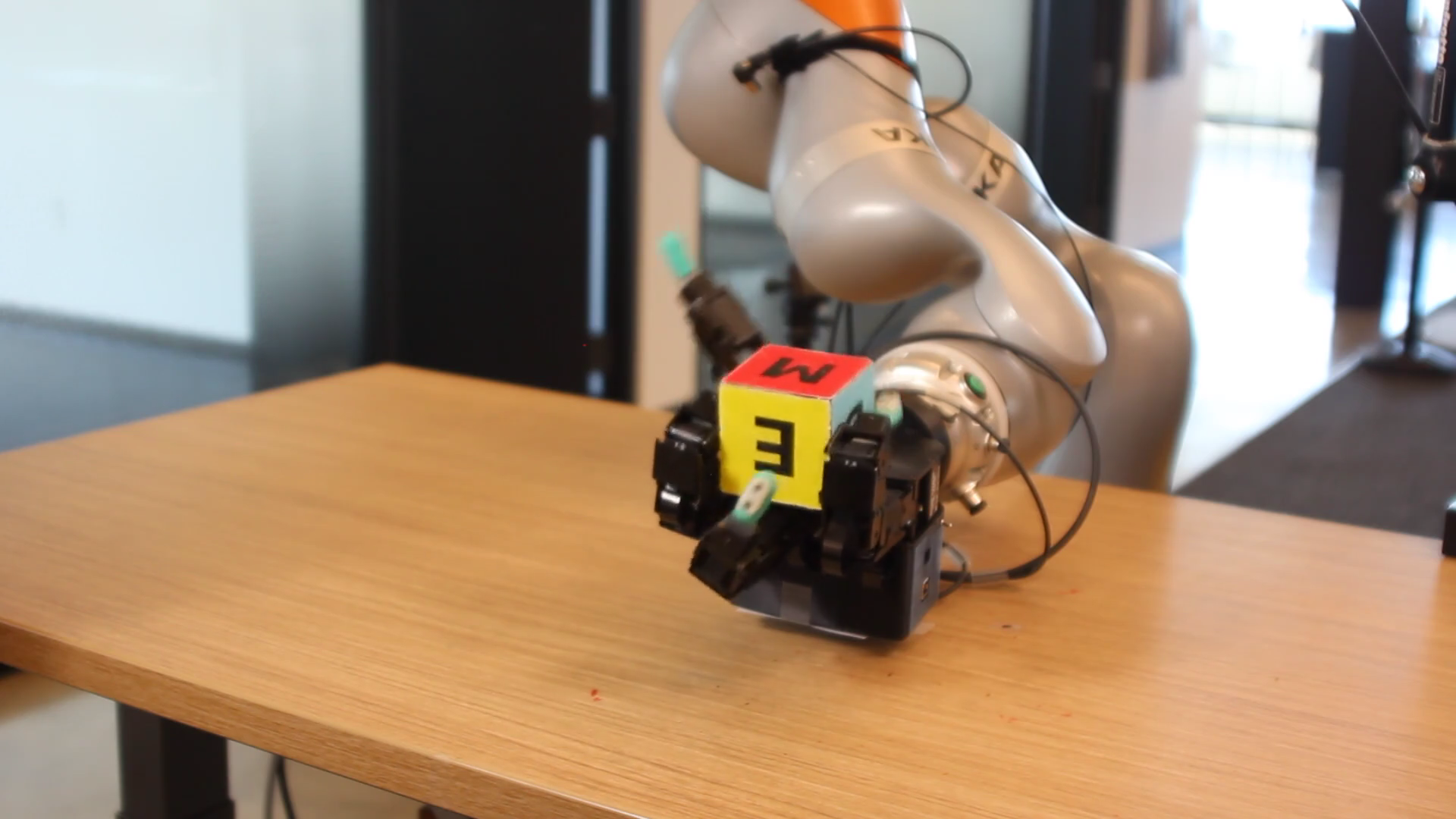}
\includegraphics[width=0.245\linewidth]{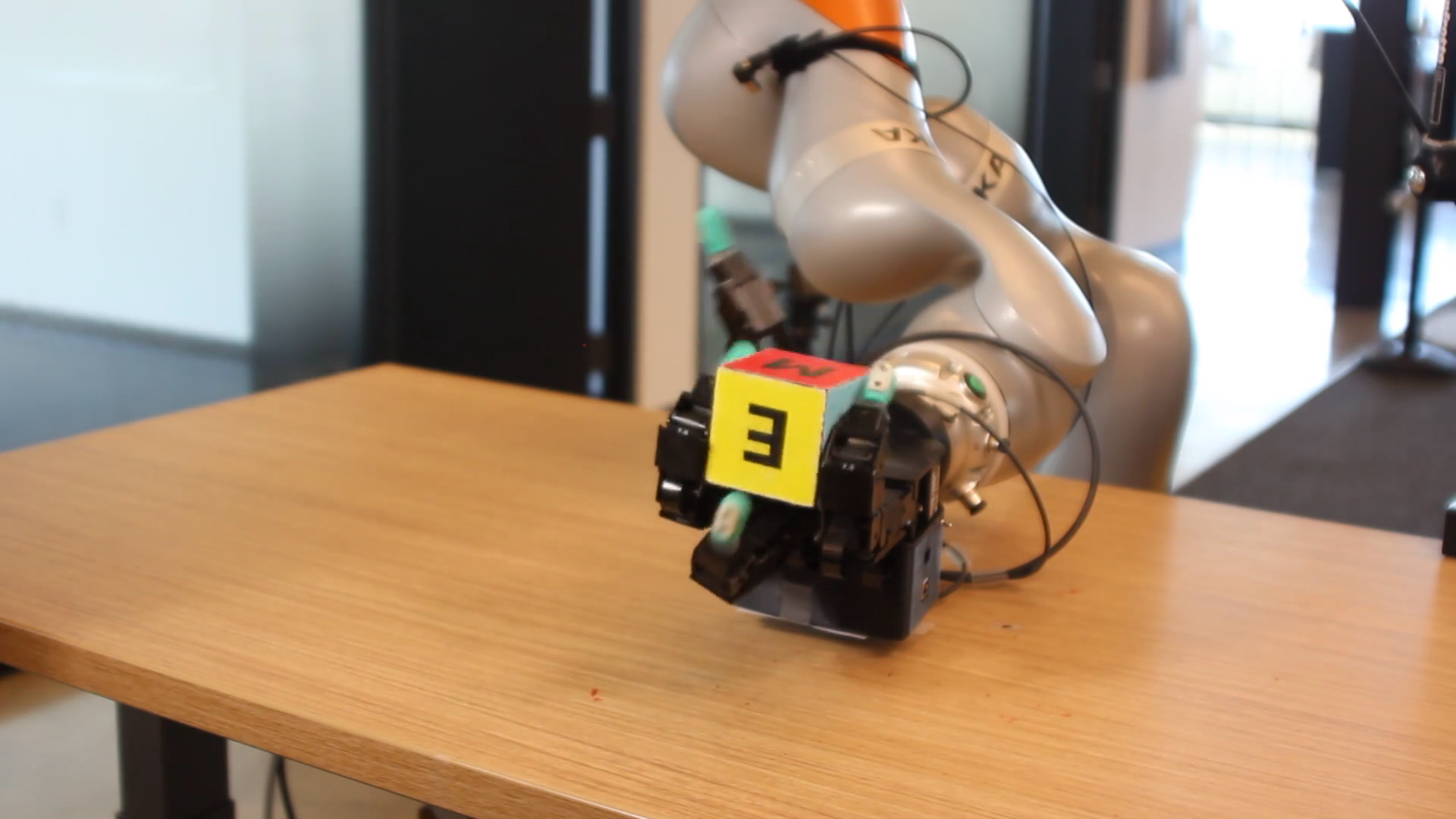}
\centerline{
\hfill
\makebox[0.24\linewidth][c] { \footnotesize{(a) $t=1.0s$}}
\hfill
\makebox[0.24\linewidth][c] { \footnotesize{(b) $t=5.0s$}}
\hfill
\makebox[0.24\linewidth][c] { \footnotesize{(c) $t=9.0s$}}
\hfill
\makebox[0.24\linewidth][c] { \footnotesize{(d) $t=12.0s$}}
}
\caption{\small
Policies trained with manual DR exhibited `stuck' behaviours, where the cube remained stuck in certain configurations and was unable to recover. An example of such behaviour can be observed here: \url{https://www.youtube.com/watch?v=tJgq18VbL3k
}.} \vspace{1mm}
\label{fig:manual_dr_stuck}
\end{figure}

We benchmark both ADR and non-ADR (manually-tuned DR ranges) policies in Table \ref{table:real_results} and like \cite{openai-sh} find that the policies trained with ADR perform the best, suggesting that the sheer diversity of data gleaned from the simulator endows the policies with the extreme robustness needed in the real world. Importantly, we observed that policies trained with non-ADR exhibited `stuck' behaviours (as shown in Figure \ref{fig:manual_dr_stuck}), which ADR-based policies were able to overcome due to increased diversity in training data. We also find that on an average, the trials with ADR achieve more consecutive successes than non-ADR policies. Table \ref{table:cs_comparison} puts our results in perspective alongside the previous works of \cite{openai-sh} and \cite{openai-rubiks}. \textbf{We demonstrate performance which significantly improves upon the best vision policies from \citep{openai-sh} and ADR (XL)\footnote{XL and XXL denote the degree of randomisations.} policies given high-quality state information from a motion capture system in \citep{openai-rubiks}}. Our policies do not achieve the average successes seen in \cite{openai-rubiks} with ADR (XXL) with \textit{state information}. We hypothesise that this maybe due to (a) better accuracy of the state information from the PhaseSpace markers (b) higher frame-rate of state observations with PhaseSpace markers (c) increased diversity of data with ADR (XXL). However, our best vision-based policy generated $\sim$2.5$\times$ higher peak consecutive successes and $\sim$1.5$\times$ higher mean consecutive successes than the vision-based policies in \cite{openai-sh} as shown in Table \ref{table:cs_comparison}.

\newlength{\oldintextsep}
\setlength{\oldintextsep}{\intextsep}
\setlength\intextsep{0pt}\begin{wraptable}{r}{5.1cm}
\begin{tabular}{c|c} 
        \toprule
        \rowcolor[HTML]{D4F7EE}
\footnotesize{Frame Hold (N)} & \footnotesize{Cons. Successes} \\\midrule
\rowcolor[HTML]{EFEFEF}
\footnotesize{0} & \footnotesize{38.4}\\  \midrule
\footnotesize{5} & \footnotesize{35.3}\\  \midrule
\rowcolor[HTML]{EFEFEF}
\footnotesize{10} & \footnotesize{33.3}\\  \midrule
\footnotesize{20} & \footnotesize{27.3} \\ \bottomrule
\end{tabular}
\caption{Performance in simulation with ADR policies with respect to $N$.}\label{table:goal_frame_hold_exp}
\vspace{-3pt}
\end{wraptable} 
Additionally, we also benchmark our results beyond the basic experiment of goal reaching by making the policy hold the cube at a target orientation $N$ frames in a row (we use $N=10$), \textit{i.e.}, we count the number of times the cube orientation is within the threshold of 0.4 rad in a row (as described in \ref{sec:training_in_sim}), refresh the goal only if the frame counter reaches $N$, and reset the counter to zero every time it goes outside the threshold. In the basic experiment of goal reaching without the hold, the cube may shoot past the target, making it difficult to tell if the target was achieved merely due to noise in the pose estimation or if the cube orientation was indeed accurately estimated. Therefore, holding the cube for $N$ frames in a row ensures that the goal was not achieved by chance, highlighting the robustness of the pose estimator. We conduct this experiment only with our ADR policies as shown in (Table \ref{table:real_results}, 2$^{nd}$ row). It is worth noting that the policy was not trained explicitly to hold the cube and that the experiment is meant to be a test of accuracy of pose\footnotetext{To fully hold the cube stationary in hand for a target orientation means zero velocities at the target, which requires changing the reward function.}. We chose $N=10$ based on our simulation experiments and found that setting $N$ too high led to a dramatic drop in the performance because the LSTM was not trained for such scenarios (see Table \ref{table:goal_frame_hold_exp}). On the other hand, setting it too low did not change the simulation performance; $N=10$ was a good balance between performance and LSTM stability. This also lets us separate the drop in performance due to LSTM instability from pose estimation errors in the real world. From the trials, we find that while there is a noticeable drop in performance \textit{i.e. }the maximum consecutive successes are only 45 as opposed to 112 in the other case, the average consecutive successes have not dropped as dramatically suggesting both the robustness of the policy and the pose estimator.

\begin{table}
\centering
\resizebox{\linewidth}{!}{
\begin{tabular}{l|c|c|c|c|c|c} 
        \toprule
        \rowcolor[HTML]{D4F7EE}
        Method & DR type & Pose estimation type & Training time & Cons. Successes & Median & Best rollout\\
        \midrule
        \rowcolor[HTML]{EFEFEF} 
        OpenAI \textit{et al.} \cite{openai-sh} (state) & Manual & PhaseSpace & 2.08 days (50 hours) & 18.8 ± 5.4 & 13.0 & 50\\
        OpenAI \textit{et al.} \cite{openai-sh} (state, locked wrist) & Manual & PhaseSpace & 2.08 days (50 hours) & 26.4 ± 7.3 & 28.5 & 50\\
        \rowcolor[HTML]{EFEFEF} 
        OpenAI \textit{et al.} \cite{openai-sh} (vision) & Manual & Neural Network & 2.08 days (50 hours) & 15.2 ± 7.8 & 11.5 & 46\\ 
        OpenAI \textit{et al.} \cite{openai-rubiks} (state) & ADR (L) & PhaseSpace & 13.76 days & 13.3 ± 3.6  & 11.5 & -- \\ 
        \rowcolor[HTML]{EFEFEF} 
        OpenAI \textit{et al.} \cite{openai-rubiks} (state) & ADR (XL) & PhaseSpace & Multiple Months & 16.0 ± 4.0 & 12.5 & --\\ 
        OpenAI \textit{et al.} \cite{openai-rubiks} (state) & ADR (XXL) & PhaseSpace & Multiple Months & 32 ± 6.4 & 42.0 & --\\
        \rowcolor[HTML]{EFEFEF} 
        Ours (vision)  & Manual & Neural Network & 1.41 days (34 hours) & 14.8 ± 5.4 & 14.5 & 29\\
        Ours (vision, best avg)  & ADR & Neural Network & 2.5 days (60 hours) & 27.8 ± 19.0 & 14.0 & 112\\ 
        \rowcolor[HTML]{EFEFEF} 
        Ours (vision, best median) & ADR & Neural Network & 2.5 days (60 hours) & 26.6 ± 13.2 & 20.5 & 77\\ 
        Ours (vision, max successes capped at 50) & ADR & Neural Network & 2.5 days (60 hours) & 23.1 ± 9.4 & 20.5 & 50\\ 
       
        \bottomrule
\end{tabular}
}
\vspace{2pt}
\caption{We compare our block reorientation results against the previous work of OpenAI \textit{et al.} \cite{openai-sh} and OpenAI \textit{et al.} \cite{openai-rubiks}. It is important to note that we use the less capable but more affordable, 4-fingered \textbf{Allegro Hand} with a different morphology and a locked wrist, whereas they use the high-end, tendon-based \textbf{Shadow Hand} with five fingers and a movable wrist. Notwithstanding this disparity between the two robot platforms, we have tried to provide the best possible comparison in this table. Our best vision based policy on average performs better than the vision-based policy in \cite{openai-sh} and the state-based policy trained with ADR(XL) in \cite{openai-rubiks} with PhaseSpace markers while taking only 2.5 days to train. OpenAI \textit{et al.} \cite{openai-rubiks} do not mention the precise time to train with ADR (XL) and ADR (XXL) but do say ``multiple months'' in the paper. Missing entries in the last column mean that the information was not available in the paper. The table only considers the time taken by the control policy to train, and training time for vision models is not included. Our best rollout with vision achieves 112 consecutive successes, which is roughly 2.5$\times$ more than theirs with vision\cite{openai-sh}. It is worth reminding that both \cite{openai-sh} and \cite{openai-rubiks} capped the maximum consecutive successes to 50 (Section 6.2, OpenAI \textit{et al.}\cite{openai-sh}). Therefore, we also provide the statistics with successes capped at 50 in the last row of this table. The compute budget comparisons are in Section \ref{sec:intro} and Appendix \ref{sec:compute_comparisons}.}
\label{table:cs_comparison}
\end{table}

\subsection{Quirks, Problems, and Surprises}
\label{sec:quirks}

\vspace{2pt}
\begin{figure}[h]
 \begin{minipage}[c]{\textwidth}
    \centering
    \includegraphics[width=0.245\linewidth]{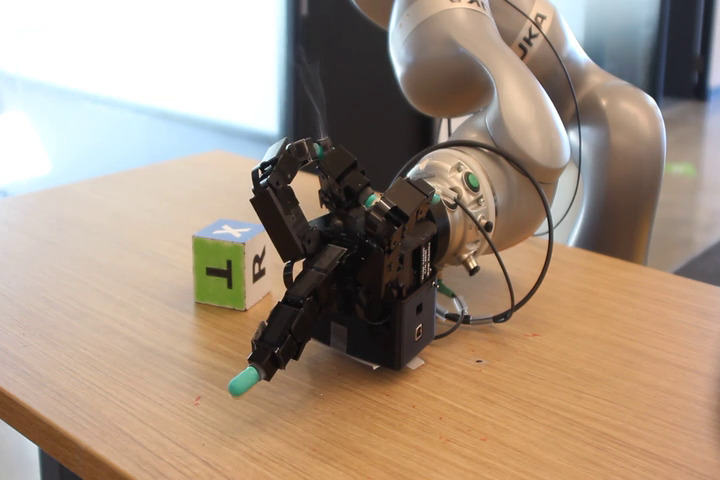}
    \includegraphics[width=0.245\linewidth]{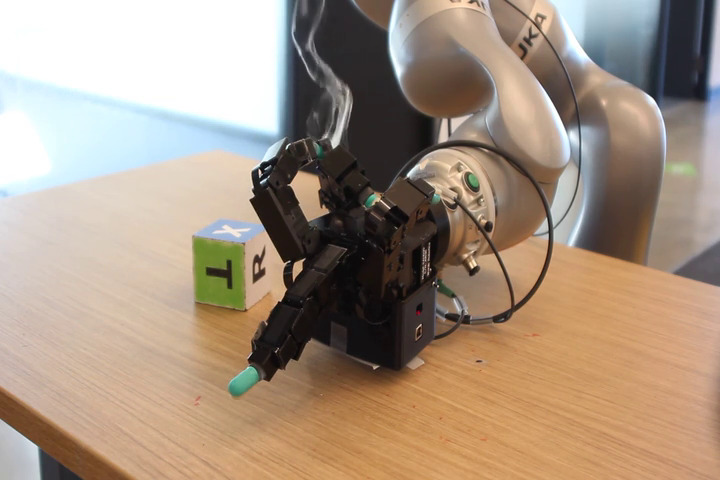}
    \includegraphics[width=0.245\linewidth]{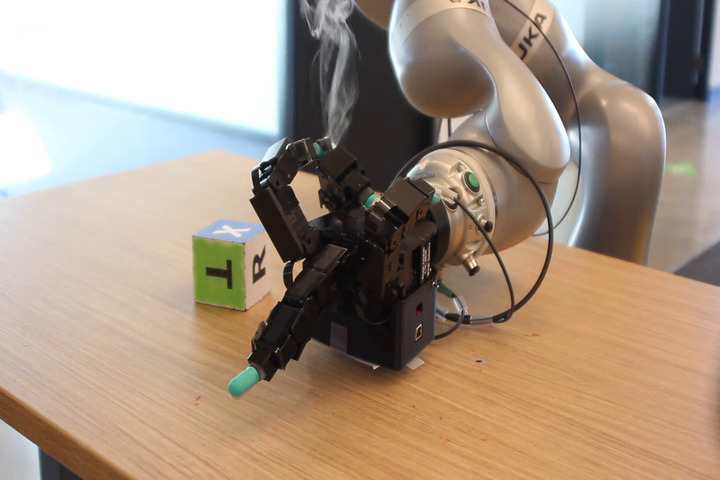}
    \includegraphics[width=0.245\linewidth]{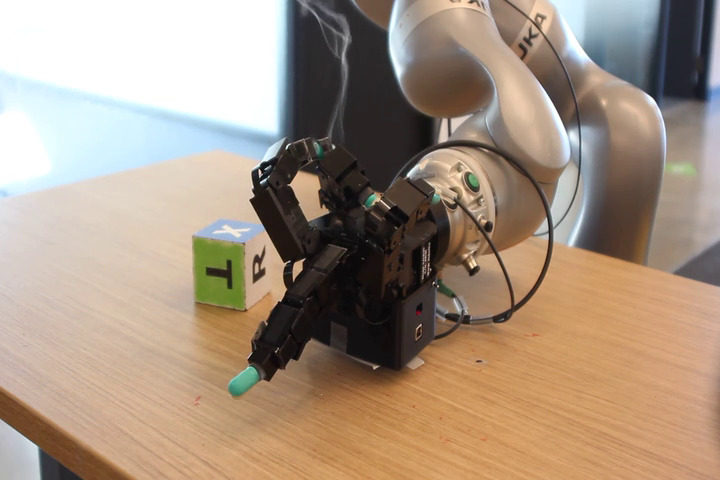}
    \captionof{figure}{Ribbon cables can burn with aggressive maneuvers or excessive usage. Full video is available at \url{https://www.youtube.com/watch?v=H5xOer9jxJc}. The images show one of the instances of running the policy leading to burning cables and smoke coming out of the hand. Regular health checkups are necessary, involving replacing cables, tightening screws and resting the hand for a few hours after running about 10-15 trials.}
    \vspace{2pt}
    \label{fig:dextreme_smoke}
\end{minipage}
\end{figure}

Throughout our work, we experienced a variety of recurring problems and observations, which we disclose here in the interest of transparency and providing avenues for future research. Due to complexities involved in training and slow turnaround time in results because of real hardware involved in the loop, we do not have rigorous experiments to prove these. However, they constitute "tricks of the trade" that will hopefully be of use to other researchers.

\begin{itemize}
    \item Even though we are not training on the real robot hand and our policies are very gentle due to the low-pass filters we apply, the hand is quite fragile and requires regular health checks and cable replacements \textit{e.g.} the ribbon cables connecting joints on the fingers can break or burn, disabling the corresponding joints (see Figure \ref{fig:dextreme_smoke}). We also apply hot glue at the either ends of the ribbon cables going into connectors to tighten the connection and prevent them from coming out.
    \item We had applied 300lse tape on fingers and palm of the hand for another unrelated task of grasping, but it added significant friction, preventing the cube from sliding and moving smoothly on the palm. Higher friction also meant that the fingers tried aggressively to manipulate the cube when it was stuck in the hand. Therefore, we removed the tape from the palm. However, we kept the tape on the sides of the fingers to allow better grips on the object.
    \item We found that deployment in the real world was quite variable between policies trained with different seeds, even with the same domain randomisation parameters. This is an issue that was also noted in \citep{openai-dynamics}. We suspect that this is because, despite the extreme levels of randomisation we do, there is a "null space" of possible policies which perform similarly in simulation but differently in the real world.
    \item Automatic Domain Randomisation \citep{openai-rubiks}, which was used to achieve our best results, does not model the joint distribution and relationships between performance with different parameters. This means it can provide different results depending on the order in which randomisation ranges increase, and furthermore it may not explore the Pareto-limits of performance on any particular dimension. This can lead to cases where ADR disproportionately increases the performance of the policy on dimension A at the expense of dimension B.

    \item Our best policy trained in a non-ADR setting exhibited slightly more natural and smooth behaviour than the ADR one. We also observed that the ADR policies tended to cage the object, resulting in letters occluded by the fingers from all cameras. This made the pose estimation unreliable in those configurations (although the policy was able to absorb the errors and performed better overall). It is possible temporal smoothing of the pose estimator can help as long as it does not introduce any latency.
    \item We also tried pose estimation without the letters on the faces and found that to be quite challenging --- the network was confused by colour changes due to lighting in the real scene and the quality of the cameras used \textit{e.g.} yellow turning into white in the image when the normal of a cube face is perpendicular to the camera. Having a robot cage as used in \cite{openai-sh, openai-rubiks} and \cite{isaacgym-trifinger} with an constrained artificial light can prevent this from happening to an extent, but it also makes the system confined to the cage.
\end{itemize}

On the other hand, we were surprised positively about some aspects of our system:

\begin{itemize}
    \item Our pose estimator proved to be surprisingly robust even in real-world scenarios outside of Allegro Hand manipulation. This hints at the power of extreme randomisation with synthetic data in simulation for developing such general systems.
    \item The ability to, at test time, adjust the speed of the policy by tuning the EMA value (see Section \ref{sec:policy-learning} - we trained with 0.15 but tested with 0.1) was very useful to avoid damaging the hardware while at the same time having agile policies.
    \item The agility of our policies in the real world on the Allegro Hand, which we initially expected to be a limitation as it is relatively large and is motor- rather than tendon-driven, was a positive surprise to us.
    \item Most of our experiments and trials conducted on the robot were done with a worn-out thumb on the Allegro Hand --- one of the wires connecting a joint to the circuit board had a loose connection --- and we were quite surprised that the policies produced high consecutive successes despite a malfunctioning actuator, suggesting the robustness of learning-based approaches. The slow turnaround time involved in repairing the hardware motivated us to do it ourselves regularly during the experiments, but it was only a temporary solution.
    \item Since we do not regress to the pose via an end-to-end network, we found that our pose estimator was not tied to a particular camera configuration as keypoint detection worked reliably from different configurations\footnote{While extrinsics change with different camera configurations, the intrinsics remain the same.} \textit{i.e.} our earlier results as described in \ref{sec:progressive_improvement} were obtained with a camera configuration that was different to the one we used afterwards with ADR. 
    \item Our best policies were trained with continuous actions unlike \citet{openai-sh,openai-rubiks}, where discrete actions were used.
\end{itemize}

\section{Related work}

In-hand manipulation is a longstanding challenge in robotics. Classical methods \cite{salisbury82, salisbury85, li89grasping} have focussed on directly leveraging the robot's kinematic and dynamic model to analytically derive controllers for the object in hand. These approaches work well while an object maintains no-slip contacts with the hand, but struggle to achieve good results in dynamic tasks where complex sequences of making-and-breaking of contacts is needed. 

Reinforcement learning has proven to be a powerful method for learning complex behaviours in robots with many degrees of freedom. \citet{Hwangbo_2019} and \citet{lee20} showed how robust legged locomotion can be learned even over challenging terrain using a combination of deep RL, fast simulation, and domain randomisation. A crucial inspiration for our work is \citet{openai-sh} on learning in-hand reorientation of cubes via reinforcement learning  in simulation and subsequent \citet{openai-rubiks} extending this task to solving Rubik's cubes. A variety of recent works \cite{isaacgym, chen21, huang21, myosuite} have leveraged new RL techniques and simulators in order to reproduce or extend in simulation the anthropomorphic in-hand manipulation capabilities shown in \cite{openai-sh}. However, these works have not shown sim-to-real transfer, demonstrating that this remains a significant challenge for learned in-hand manipulation. There have also been a variety of recent approaches attempting in-hand manipulation from the perspective of finger gaiting \cite{higo18, Morgan:etal:RAL2022}. However, these often fail to reproduce the agile dexterity present in human hands, as the limitations of such a sequential approach to control place corresponding limits on speed. Similarly, \citet{isaacgym-trifinger} and \citet{shi2020circus} achieve real-world in hand manipulation using reinforcement learning, but on a platform that cannot mimic the dexterity of a human hand. \citet{dlrhand22} learned in-hand manipulation using tactile sensing, but the lack of vision meant that the kinds of re-posing behaviour and speed of the manipulation were limited. There has also been research investigating using reinforcement learning directly on hardware platforms as opposed to in simulation \cite{vikashCoffeeBeans, kumarImitation}. However, this limits the complexity of learning that can be done due to the small number of trials available on real hardware platforms, as well as the wear-and-tear imposed.

Pose estimation for robotic manipulation is a widely studied area \cite{tremblay2018deep, cosypose}. However, relatively few works outside of \citet{openai-sh} have applied it to the problem of contact-rich, dexterous, in-hand manipulation, which introduces challenges that exclude many off-the-shelf pose estimators (due to the large degree of occlusions, motion blur, \textit{etc.}).

\section{Limitations}
\label{sec:limitations}
Despite our best efforts, the gap between simulations and the real world is still noticeable. Our non-ADR (manual DR) based policy achieves an average of 35 consecutive successes (see Figure \ref{fig:training-curves}(a)) in simulation, but only obtains an average of about 15 consecutive successes in the real world. While our ADR policies do perform better, they still fail to reach the average successes seen in simulations. Another limitation of our work is that our ADR policies do not consistently obtain high consecutive successes as seen in ADR (XXL) in \citet{openai-rubiks}. It is unclear whether we need a more reliable pose estimation \textit{e.g.} marker-based systems, the policy needs more diversity in the data, or if it was due to the malfunctioning thumb on our Allegro Hand. Our best npd with ADR policies is around -0.2, and it is possible that further improvements can be made. This is something we would like to investigate in future work.

We also observed that there is still some sim-to-real gap in pose estimation. This is manifested when we played back the real states in sim (real-to-sim) with physics enabled, which sometimes resulted in interpenetrations. Therefore, we were not able to easily calibrate physics parameters of the cube, which could have improved our sim-to-real transfer.   

Lastly, putting this work in context, it is worth remembering that some of the key reasons for successful transfer of this task involve: 
\begin{itemize}
    \item Being able to simulate the task, and having a clear and well-defined reward function so that the policies can be trained in simulation.
    \item Randomisation of interpretable parameters exposed by the simulator and providing a curriculum for training in simulation. For instance, we randomised \textit{friction, damping, stiffness, etc.}, which were crucial for the transfer. ADR provided a curriculum.
    \item Ability to evaluate successful execution of the task so that we can track the improvements. In this work, we compared the current orientation with the desired, and if the difference was within a user-specified threshold (0.4 rad in our case), the goal-reaching was considered successful.
\end{itemize}

Many real-world tasks are hard to simulate and sometimes defining reward functions is not possible as it is in this task. Even if we could simulate and have a well-defined reward function, evaluating a successful execution of a task, \textit{e.g.} cooking a meal, is not straightforward.

\section{Acknowledgements}
The authors would like to thank Nathan Ratliff, Lucas Manuelli, Erwin Coumans, and Ryan Hickman for helpful discussions. Maciej Bala assisted with multi-GPU training; Michael Lin helped with hardware and Zhutian Yang with video editing. Nick Walker provided valuable suggestions on the teaser video and proofreading. Thanks also to Ankit Goyal for the help with high frame-rate video capture and Jie Xu for proofreading.

\newpage
\section{Contributions}
\label{sec:contributions}

\textbf{Reinforcement Learning \& Simulation}

\begin{itemize}
    \item \textbf{Viktor Makoviychuk} implemented the first version of the Allegro Hand environment in Isaac Gym. 
    \item \textbf{Ankur Handa, Arthur Allshire, and Viktor Makoviychuk} developed the domain randomisations in Isaac Gym that assisted in sim2real transfer.
    \item \textbf{Arthur Allshire} developed the vectorised Automatic Domain Randomisation (ADR) in Isaac Gym.
    \item \textbf{Ankur Handa, Arthur Allshire, Viktor Makoviychuk, and Aleksei Petrenko} trained RL policies and added various features to the environment to improve sim2real transfer.
    \item \textbf{Denys Makoviichuk} implemented the RL Library used in this project and helped implement specific extensions required for this work. 
    \item \textbf{Yashraj Narang and Gavriel State} advised on tuning simulations.
    \item \textbf{Dieter Fox and Gavriel State} advised on experiments.
\end{itemize}

\textbf{Vision}

\begin{itemize}
    \item \textbf{Ankur Handa} developed the code to train pose estimation and data augmentation for the vision models.
    \item \textbf{Arthur Allshire} wrote the vision data rendering system and developed the domain randomisations in Omniverse. \textbf{Ritvik Singh and Jingzhou Liu} generalised and extended the rendering pipeline.
    \item \textbf{Ankur Handa, Ritvik Singh, and Jingzhou Liu} trained pose estimation models and did real-to-sim with vision to improve the pose estimator.
    \item \textbf{Alexander Zhurkevich} helped speed up the inference.
\end{itemize}

\textbf{Real-World Experiments}

\begin{itemize}
    \item \textbf{Ankur Handa} conducted the real-world experiments. \textbf{Arthur Allshire} helped with experiments in the early stages, and \textbf{Viktor Makoviychuk} helped with experiments in the later stages of the project.
    \item \textbf{Arthur Allshire} wrote the code to perform policy \& vision inference on the real robot. \textbf{Ankur Handa} maintained and extended it.
    \item \textbf{Arthur Allshire} developed the live visualisation pipeline in Omniverse.
    \item \textbf{Karl Van Wyk} managed the Allegro Hand infrastructure to run experiments on the real hand. 
    \item \textbf{Balakumar Sundaralingam and Ankur Handa} provided assistance repairing the Allegro Hand when it broke.
    \item \textbf{Karl Van Wyk} developed an automatic calibration system for camera-camera and camera-robot calibration.
\end{itemize}

\textbf{Organisational}

\begin{itemize}
    \item \textbf{Arthur Allshire and Ankur Handa} drafted the paper.
    \item \textbf{Yashraj Narang, Ritvik Singh, Jingzhou Liu, and Gavriel State} helped to edit the paper.
    \item \textbf{Gavriel State, Dieter Fox, and Jean-Francois Lafleche} provided resources and support for the project.
    \item \textbf{Ankur Handa} edited the videos. \textbf{Gavriel State, Arthur Allshire, Viktor Makoviychuk, Jingzhou Liu, Yashraj Narang and Ritvik Singh} examined the videos and provided feedback.
    \item \textbf{Ankur Handa} led the project. \textbf{Ankur Handa, Arthur Allshire} and \textbf{Viktor Makoviychuk} designed the roadmap of the project.
\end{itemize}

\newpage

\clearpage
\newpage

\bibliographystyle{unsrtnat}
\bibliography{ig-report}

\newpage

\appendix

\section{Appendix}

\subsection{Compute Budget Comparisons}
\label{sec:compute_comparisons}

\begin{table}[htp]
\centering
\resizebox{\linewidth}{!}{
\begin{tabular}{c|c|c|c|c} 
        \toprule
        \rowcolor[HTML]{D4F7EE}
        Method & DR type & Compute Infrastructure & Training time & Training Cost\\
        \midrule
        \rowcolor[HTML]{EFEFEF} 
        OpenAI \textit{et al.} \cite{openai-sh} & Manual DR & 384 CPU servers with 16-cores each and 8 NVIDIA V100s & 2.08 days & \$14,280.0\\
        OpenAI \textit{et al.} \cite{openai-rubiks} & ADR & 400 CPU servers with 32-cores each and 32 NVIDIA V100s & 13.76 days & \$215,685.1\\ 
        \rowcolor[HTML]{EFEFEF} 
        Ours & Manual DR & 8 NVIDIA A40s & 1.41 days & \$553.8\\ 
        Ours & ADR & 8 NVIDIA A40s & 2.50 days & \$977.2\\ 
        \bottomrule
\end{tabular}
}
\vspace{2pt}
\caption{Compute budget comparisons for block reorientation task of our work against the previous work.}
\label{table:compute_budget_requirements}
\end{table}

The costs are estimated from the AWS EC2 instance pricing as of $October$ 19, 2022 at \url{https://ec2pricing.net/}. The estimated costs for \citet{openai-sh, openai-rubiks} may have been higher in 2018 and 2019 respectively.

For the OpenAI \cite{openai-sh} equivalent, a \texttt{p3.16xlarge} 8$\times$V100 machine at \$24.48/hr and a \texttt{c6i.4xlarge} with 16 CPU cores at \$0.68/hr add up to a total cost of \$(24.48 + 0.68$\times$384)$\times$50=\$14280. Similarly, for an equivalent of OpenAI \cite{openai-rubiks} with ADR, a 32-core \texttt{c6i.8xlarge} machine at \$1.36/hr, the overall cost can be estimated to be \$(24.48$\times$4 + 1.36$\times$400)$\times$24$\times$14=\$215,685. 

While AWS does not offer NVIDIA A40s, it does provide NVIDIA A10Gs which are similar in performance. A \texttt{g5.48xlarge} instance with 8 NVIDIA A10G GPUs costs at \$16.288/hr. For our manual DR experiments the overall compute adds up to \$16.288$\times$34=\$553.8 and for ADR it is \$16.288$\times$60=\$977.28 in total. The cost for such experiments on \texttt{p4d.24xlarge} instances offering 8-GPU A100 machines at \$32.77/hr would be \$1114.18 and \$1966.2 respectively.

\subsection{Hardware Comparisons}
\label{sec:hardware_comparisons}

\begin{table}[htp]
\centering
\resizebox{\linewidth}{!}{
\begin{tabular}{c|c|c|c|c|c|c } 
        \toprule
        \rowcolor[HTML]{D4F7EE}
        Method & Robot Hand & RGB & Markers on Cube & Markers on Fingertips & Tactile Sensors & Robot Cage\\
        \midrule
        \rowcolor[HTML]{EFEFEF} 
        OpenAI \textit{et al.} \cite{openai-sh} & Shadow Hand & \checkmark  & \checkmark & \checkmark & \ding{55} & \checkmark\\
        OpenAI \textit{et al.} \cite{openai-rubiks} & Shadow Hand & \ding{55} & \checkmark & \checkmark & \ding{55} & \checkmark\\ 
        \rowcolor[HTML]{EFEFEF} 
        Ours & Allegro Hand & \checkmark  & \ding{55} & \ding{55} & \ding{55} & \ding{55} \\
        \bottomrule
\end{tabular}
}
\vspace{2pt}
\caption{Our hardware setup compared against the one used in \citet{openai-sh} and \citet{openai-rubiks}. Note that the experiment pertaining to the block reorientation in \cite{openai-rubiks} was done with marker-based systems and not RGB cameras.}
\label{table:hardware_comparisons}
\end{table}

\subsection{PPO Hyperparameters}
\label{sec:ppo_hyperparams}
\vspace{2pt}
\begin{table}[htp]
\centering
\resizebox{0.6\linewidth}{!}{
\begin{tabular}{c|c} 
        \toprule
        \rowcolor[HTML]{D4F7EE}
        Hyperparameter & Value\\
        \midrule
        \rowcolor[HTML]{EFEFEF} 
        Hardware configuration & 8 NVIDIA A40\\
        Action distribution & 16D Continuous actions\\
        \rowcolor[HTML]{EFEFEF} 
        Discount factor & 0.998\\
        Generalised advantage estimation & 0.95\\
        \rowcolor[HTML]{EFEFEF} 
        Entropy regularisation coefficient & 0.002\\
        PPO clipping parameter & 0.2 \\
        \rowcolor[HTML]{EFEFEF} 
        KL-divergence threshold  & 0.16\\
        Optimiser & Adam \\
        \rowcolor[HTML]{EFEFEF} 
        Learning rate (Actor) &  1e-4 \\
        Learning rate (Critic) &  5e-4 \\
        \rowcolor[HTML]{EFEFEF} 
        Minibatch size & 16384\\
        Learning epochs & 4 \\
        \rowcolor[HTML]{EFEFEF} 
        Horizon length & 16 \\
        LSTM input sequence length & 16 \\ 
        \rowcolor[HTML]{EFEFEF} 
        Action bounds loss coefficient & 0.005 \\

        \bottomrule
\end{tabular}
}
\vspace{2pt}
\caption{Various hyperparameters used to train the policy with PPO.}
\label{table:ppo_params}
\end{table}

\newpage

\subsection{Isaac Gym Simulation Parameters}
\label{sec:ig_hyperparams}
\vspace{2pt}
\begin{table}[!htbp]
\centering
\resizebox{0.4\linewidth}{!}{
\begin{tabular}{c|c} 
        \toprule
        \rowcolor[HTML]{D4F7EE}
        Hyperparameter & Value\\
        \midrule
        \rowcolor[HTML]{EFEFEF} 
        Sim $dt$ & $1/60s$\\
        Control $dt$ & $1/30s$\\
        \rowcolor[HTML]{EFEFEF}
        Num Envs & 8192/16384\\
        Num Substeps & 2\\ 
        \rowcolor[HTML]{EFEFEF}
        Num Pos Iterations & 8\\ 
        Num Vel Iterations & 0\\ 
        \bottomrule
\end{tabular}
}
\vspace{2pt}
\caption{Simulation settings for the Allegro Hand environment.}
\label{table:ig_params}
\end{table}

\subsection{Progressive Improvements in Consecutive Successes in the Real World}
\label{sec:progressive_improvement}
\vspace{2pt}
\begin{table}[htp]
\centering
\resizebox{0.8\linewidth}{!}{
\begin{tabular}{c|c|c} 
        \toprule
        \rowcolor[HTML]{D4F7EE}
        Date & Max Consecutive Success & Notes\\
        \midrule
        \rowcolor[HTML]{EFEFEF} 
        29-Mar-2022  & 10 & Manual DR  \\
        14-Jun-2022   & 15 & Pose wrt wrist\\
        \rowcolor[HTML]{EFEFEF}
        21-Jun-2022 & 20 & Increased DR ranges\\
        26-Jul-2022 & 31 & Increased network sizes \\ 
        \rowcolor[HTML]{EFEFEF}
        27-Jul-2022  & 43 & Random pose injection and RNA  \\ 
        20-Aug-2022   & 70 & ADR\\ 
        \rowcolor[HTML]{EFEFEF}
        27-Aug-2022   & 77 & ADR \\ 
        30-Aug-2022   & 112 & ADR and deeper networks \\ 
        \bottomrule
\end{tabular}
}
\vspace{2pt}
\label{table:progressive_cs_improv}
\end{table}

\subsection{Default KUKA configuration}

\begin{table}[!htp]
\centering
\resizebox{0.9\linewidth}{!}{
\begin{tabular}{c|c|c|c|c|c|c|c}
\toprule
\rowcolor[HTML]{D4F7EE}
Joint Name & A1     & A2     & A3    & A4     & A5   & A6    & A7    \\ \hline
           & -42.38 & -53.38 & 66.01 & 116.45 & 7.20 & 57.55 & 81.53 \\ 
\bottomrule
\end{tabular}
}
\vspace{2pt}
\caption{Default rest configuration of the KUKA arm joints (in degrees) used in this work.}
\end{table}

\subsection{Software Tools Used in the Work}
\vspace{2pt}
\begin{table}[!htp]
\centering
\resizebox{0.9\linewidth}{!}{
\begin{tabular}{c|c} 
        \toprule
        \rowcolor[HTML]{D4F7EE}
        Software & Source\\
        \midrule
        \rowcolor[HTML]{EFEFEF} 
        Numpy \cite{Harris2020array} & \url{https://numpy.org/}\\
        Scipy \cite{2020SciPy-NMeth} & \url{https://scipy.org/}\\
        \rowcolor[HTML]{EFEFEF}
        Matplotlib \cite{Hunter:2007} & \url{https://matplotlib.org/}\\
        ROS \cite{Quigley09} & \url{http://wiki.ros.org/rospy}\\ 
        \rowcolor[HTML]{EFEFEF}
        Isaac Gym \cite{isaacgym} & \url{https://developer.nvidia.com/isaac-gym}\\ 
        IsaacGymEnvs \cite{isaacgym} & \url{https://github.com/NVIDIA-Omniverse/IsaacGymEnvs} \\ 
        \rowcolor[HTML]{EFEFEF}
        RL Games \cite{rl-games2022} & \url{https://github.com/Denys88/rl_games} \\ 
        PyTorch \cite{PyTorch2019NeurIPS} & \url{https://pytorch.org/}\\ 
        \rowcolor[HTML]{EFEFEF}
        PyTorch3D \cite{Ravi2020pytorch3d} & \url{https://pytorch3d.org/}\\
        Roma \cite{roma:arxiv21} & \url{https://github.com/naver/roma} \\ 
        \rowcolor[HTML]{EFEFEF}
        Pangolin & \url{https://github.com/stevenlovegrove/Pangolin} \\ 
        Omniverse Isaac Sim & \url{https://developer.nvidia.com/isaac-sim} \\ 
        \rowcolor[HTML]{EFEFEF}
        iMovie & \url{https://www.apple.com/imovie/} \\ 
        \bottomrule
\end{tabular}
}
\vspace{2pt}
\label{table:software_tools}
\end{table}

\newpage
\label{dextreme-model-card}
We present a model card for DeXtreme in Table~\ref{tab:model-card}, following \citet{mitchell2019model}.

\begin{center}
\begin{longtable}{p{0.35\linewidth} | p{0.6\linewidth}}
    
    \toprule
    \noalign{\vskip 2mm}
    \multicolumn{2}{c}{\textbf{Model Details}} 
    \vspace{2mm}\\
    \toprule
    Organization Developing the Model & NVIDIA  \\
    \midrule
    Model Date & October 2022 \\
    \midrule
    Model Type & torchvision mask-rcnn inspired model for pose estimation (details in Section \ref{sec:pose-estimation}) and LSTM model for policy training (details in Section \ref{sec:policy-learning}) .
      \\
    \midrule
    Feedback on the Model &
    \href{mailto:ahanda@nvidia}{ahanda@nvidia.com}, \href{mailto:aallshire@nvidia}{aallshire@nvidia.com}, \href{mailto:jasonliu@nvidia.com}{jasonliu@nvidia.com}, \href{mailto:ritviks@nvidia}{ritviks@nvidia.com},
    \href{mailto:gstate@nvidia}{gstate@nvidia.com}, 
    \href{mailto:vmakoviychuk@nvidia}{vmakoviychuk@nvidia.com}
    \vspace{1mm} \\
    
    \toprule
    \noalign{\vskip 2mm}
    \multicolumn{2}{c}{\textbf{Intended Uses}} 
    \vspace{2mm} \\
    \toprule
    Primary Intended Uses &
    The primary use is research on pose estimation and training control policies conditioned on the pose.
    \\
    \midrule
    Primary Intended Users & Researchers working in sim-to-real. We will make some of the code available publicly.
    \\
    \midrule
    Out-of-Scope Uses &
    None. The models are meant to work on a particular combination of hardware used in this work.
    \\

    \toprule
    \noalign{\vskip 2mm}
    \multicolumn{2}{c}{\textbf{Metrics}} 
    \vspace{2mm} \\
    \toprule
    Model Performance Measures &
We focus on the performance of model in very challenging conditions \textit{e.g.} occlusions, blur and lighting changes. However, since there is no real-world dataset with ground truth for the use case we have. Therefore, most of our quantitative evaluations are done only in simulated settings Section \ref{sec:pose-estimation}.  \\
    \midrule
    Decision thresholds & N/A \\
    \midrule
    Approaches to Uncertainty and Variability &
    N/A
    \vspace{1mm} \\
    
    \toprule
    \noalign{\vskip 2mm}
    \multicolumn{2}{c}{\textbf{Training Data}} 
    \vspace{2mm} \\
    \toprule
    Datasets & We use our dataset rendered with NVIDIA Omniverse. We do not intend to make that dataset public. \\
    \midrule
    Motivation &
    We test the pose estimator in various challenging conditions. The simulation parameters are tuned to optimise the performance in the real world. The policy is trained with RL with domain randomisation to enable sim-to-real transfer.\\
    \midrule
    Preprocessing &
    Images are processed so that their mean and variance are 0 and 1 respectively. Similarly, for policy we subtract mean and divide by standard deviation.
    \vspace{1mm} \\

    \toprule
    \noalign{\vskip 2mm}
    \multicolumn{2}{c}{\textbf{Evaluation Data}} 
    \vspace{2mm} \\
    \toprule
    Datasets & We test the model in laboratory settings on real-world data collected with our robot. We do not intend to make that dataset public. \\
    \vspace{1mm} \\

    \toprule
    \noalign{\vskip 2mm}
    \multicolumn{2}{c}{\textbf{Quantitative Analyses}}
    \vspace{2mm}\\
    \toprule
    Unitary Results &
    We refer to Section~\ref{sec:real_world_perf} for the full details of our quantitative study.
    \\ 

    \toprule
    \noalign{\vskip 2mm}
    \multicolumn{2}{c}{\textbf{Performance and Limitations}} 
    \vspace{2mm} \\
    \toprule 
    \multicolumn{2}{c}{\begin{tabular}{p{\linewidth}}Section \ref{sec:real_world_perf} is about performance of the model in the real world and Section \ref{sec:limitations} describes the current limitations of our work.\end{tabular}}    
    \vspace{1mm} \\

    \bottomrule
    
    \caption{\textbf{DeXtreme Model Card.} We follow the framework presented in \citet{mitchell2019model}.}
    \label{tab:model-card}
\end{longtable}
\end{center}

\end{document}